\pdfoutput=1

\documentclass[11pt]{article}

\usepackage{acl}

\usepackage{times}
\usepackage{latexsym}
\usepackage{CJKutf8}
\usepackage{amsmath}
\usepackage{verbatim}
\usepackage{booktabs}
\usepackage{bm}
\usepackage{amssymb}
\usepackage{pifont}
\usepackage{stfloats}
\usepackage{multicol}
\usepackage{float}
\usepackage{graphicx}
\usepackage{subfigure}
\usepackage{multirow}
\usepackage{array}
\newcommand{\PreserveBackslash}[1]{\let\temp=\\#1\let\\=\temp}
\newcolumntype{C}[1]{>{\PreserveBackslash\centering}p{#1}}
\newcolumntype{R}[1]{>{\PreserveBackslash\raggedleft}p{#1}}
\newcolumntype{L}[1]{>{\PreserveBackslash\raggedright}p{#1}}
\usepackage{diagbox}
\usepackage{arydshln}
\usepackage{booktabs}
\usepackage[ruled,linesnumbered,lined,commentsnumbered]{algorithm2e}
\usepackage{mathtools}
\usepackage{amsfonts}
\usepackage{bbm}

\usepackage[T1]{fontenc}

\usepackage[utf8]{inputenc}

\usepackage{microtype}

\title{Reducing Position Bias in Simultaneous Machine Translation\\with Length-Aware Framework}

\author{Shaolei Zhang \textsuperscript{\rm 1,2},
    Yang Feng \textsuperscript{\rm 1,2}\thanks{ $\;\;$Corresponding author: Yang Feng.} \\
        \textsuperscript{\rm 1}{Key Laboratory of Intelligent Information Processing} \\ Institute of Computing Technology, Chinese Academy of Sciences (ICT/CAS) \\
    { \textsuperscript{\rm 2} {University of Chinese Academy of Sciences, Beijing, China}} \\
     \texttt{\{zhangshaolei20z, fengyang\}@ict.ac.cn}  }

\begin{document}
\maketitle
\begin{abstract}

Simultaneous machine translation (SiMT) starts translating while receiving the streaming source inputs, and hence the source sentence is always incomplete during translating. Different from the full-sentence MT using the conventional seq-to-seq architecture, SiMT often applies prefix-to-prefix architecture, which forces each target word to only align with a partial source prefix to adapt to the incomplete source in streaming inputs. However, the source words in the front positions are always illusoryly considered more important since they appear in more prefixes, resulting in \emph{position bias}, which makes the model pay more attention on the front source positions in testing. In this paper, we first analyze the phenomenon of position bias in SiMT, and develop a \emph{Length-Aware Framework} to reduce the position bias by bridging the structural gap between SiMT and full-sentence MT. Specifically, given the streaming inputs, we first predict the full-sentence length and then fill the future source position with positional encoding, thereby turning the streaming inputs into a pseudo full-sentence. The proposed framework can be integrated into most existing SiMT methods to further improve performance. Experiments on two representative SiMT methods, including the state-of-the-art adaptive policy, show that our method successfully reduces the position bias and thereby achieves better SiMT performance.

\end{abstract}

\section{Introduction}

Simultaneous machine translation (SiMT) \cite{Cho2016,gu-etal-2017-learning,ma-etal-2019-stacl,Arivazhagan2019} starts translating while receiving the streaming source inputs, which is crucial to many live scenarios, such as simultaneous interpretation, live broadcast and synchronized subtitles. Compared with full-sentence machine translation (MT) waiting for the complete source sentence, SiMT is more challenging since the source sentence is always incomplete during translating.

\begin{figure}[t]
\centering
\subfigure[Full-sentence MT with seq-to-seq architecture]{
\includegraphics[width=2.9in]{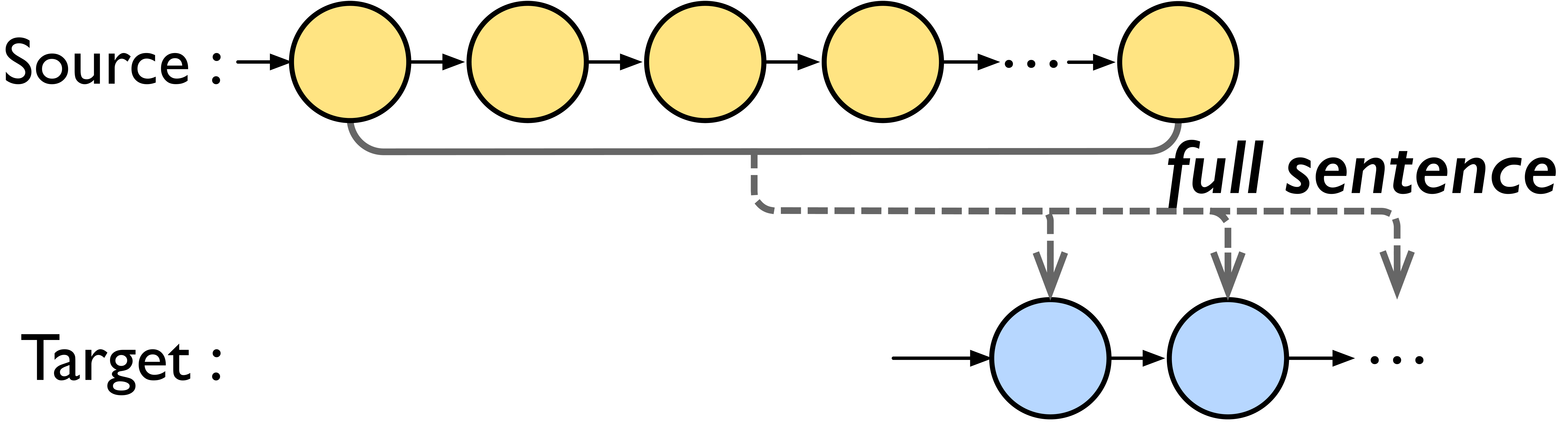}
}
\subfigure[SiMT with prefix-to-prefix architecture]{
\includegraphics[width=2.9in]{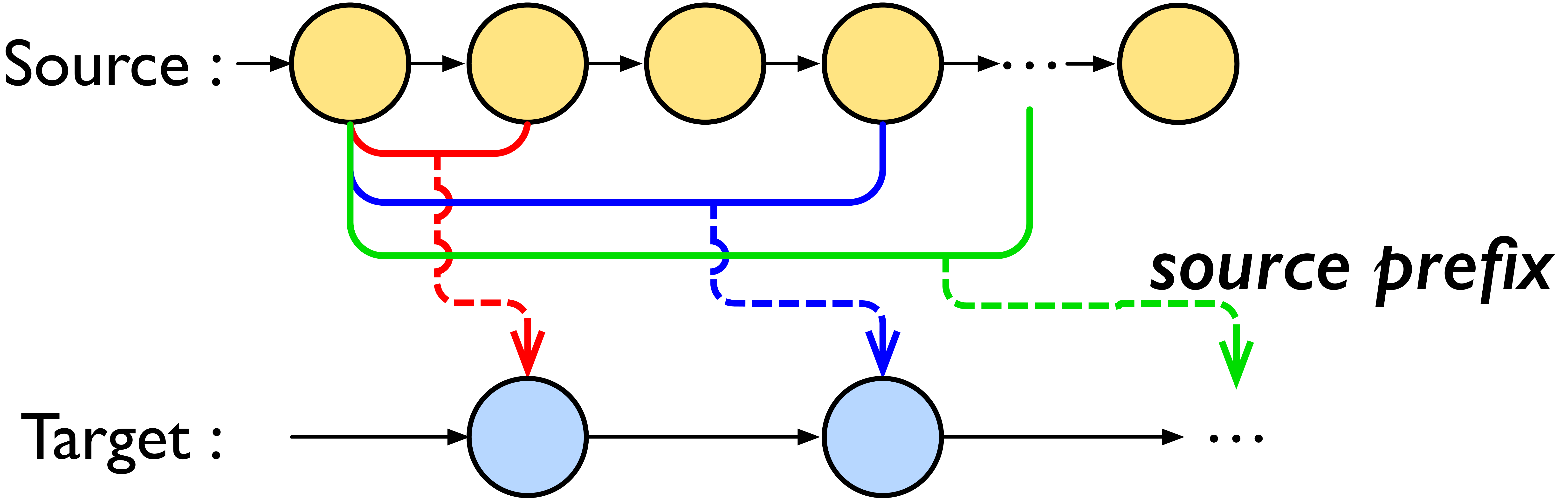}
}

\caption{Architecture of full-sentence MT and SiMT.}
\label{ill}
\end{figure}

To process the incomplete source, SiMT has a different architecture from full-sentence MT, as shown in Figure \ref{ill}. Full-sentence MT applies the \emph{seq-to-seq architecture} \cite{10.5555/2969033.2969173}, where each target word can be translated based on a complete source sentence. SiMT always applies \emph{prefix-to-prefix} architecture \cite{ma-etal-2019-stacl} to force each target word to only align with a source prefix rather than the complete source sentence, where the source prefix consists of partial source words in the front position and is monotonically non-decreasing at each step.

Although the prefix-to-prefix architecture effectively adapts to the streaming inputs by removing the subsequent source words, it intensifies the structural gap between SiMT and full-sentence MT, resulting in the following issues. First, since each target word is forced to align with a monotonically non-decreasing source prefix, the source words in different positions become no longer fair. Specifically, the source words in the front position participate in more target words’ translation due to earlier appearance, and hence are always illusoryly considered more important, resulting in \emph{position bias} \cite{ko-etal-2020-look, yan-etal-2021-position}. Due to the position bias, SiMT model prefers to pay more attention to the source words in front position during testing, which not only robs the attention of the words that are supposed to be aligned (increase mis-translation error) \cite{zhang-feng-2021-modeling-concentrated}, but also results in great overlap on attention distribution (aggravate the duplication translation error) \cite{elbayad-etal-2020-online}. We will analyze the detailed causes and disadvantages of position bias in Sec.\ref{sec:positionbias}. Second, prefix-to-prefix architecture directly removes the subsequent source words, resulting in the lost of some potential full-sentence information \cite{future-guided}. Most importantly, the prefix-to-prefix training makes the model insensitive to the full-sentence length, which can provide a global planning for translation \cite{Feng_Xie_Gu_Shao_Zhang_Yang_Yu_2020,feng-etal-2021-guiding}.

Under these grounds, we propose a \emph{Length-Aware Framework} (\emph{LAF}) for SiMT to turn the incomplete source into a pseudo full-sentence, thereby reducing the position bias. We aim to extend the incomplete source sentence in SiMT to the full-sentence length and meanwhile guarantee that future source words would not be leaked to fulfill the streaming inputs during testing. To this end, LAF first predicts the full-sentence length based on the current incomplete source sentence. Then, LAF fills the future source positions (between the current source length and predicted full-sentence length) with the positional encoding \cite{NIPS2017_7181} to construct the pseudo full-sentence. Accordingly, each target word is translated based on the pseudo full-sentence and no longer forced to align with the source prefix. LAF can be integrated into most of the existing SiMT methods to further improve performance by bridging the structural gap between SiMT and full-sentence MT.

We apply LAF on two representative and strong SiMT methods, and experiments on IWSLT15 En$\rightarrow$Vi and WMT15 De$\rightarrow$En tasks show that our method achieves better performance in both cases.

\section{Background}
We first introduce full-sentence MT and SiMT with the focus on the prefix-to-prefix architecture.

\subsection{Full-sentence Machine Translation}
For a translation task, we denote the source sentence as $\mathbf{x}\!=\!\left \{ x_{1},\cdots ,x_{J} \right \}$ with source length $J$, and target sentence as $\mathbf{y}\!=\!\left \{ y_{1},\cdots ,y_{I} \right \}$ with target length $I$. Transformer \cite{NIPS2017_7181} is the currently most widely used model for full-sentence MT, which consists of encoder and decoder. The encoder maps $\mathbf{x}$ into the source hidden states $\mathbf{h}\!=\!\left \{ h_{1},\cdots ,h_{J} \right \}$, and the decoder generates the $i^{th}$ target word $y_{i}$ based on source hidden states $\mathbf{h}$ and previous target words $y_{<i}$. Overall, the decoding probability of full-sentence MT is:
\begin{gather}
    p_{\!full}(\mathbf{y}\mid \mathbf{x})=\prod_{i=1}^{I}p\left ( y_{i}\mid \mathbf{x},\mathbf{y}_{< i} \right )
\end{gather}

\textbf{Attention} Transformer calculates the attention weights with dot-product attention, and the encoder-decoder cross-attention $\alpha _{ij}$ is calculated based on target hidden state $s_{i}$ and source hidden state $h_{j}$:
\begin{gather}
    \alpha _{ij}=\mathrm{softmax}\left ( \frac{s_{i}W^{Q}\left (h_{j}W^{K} \right )^{\top} }{\sqrt{d_{k}}} \right ) \label{eq1}
\end{gather}
where $W^{Q}$ and $W^{K}$ are input matrices, and $d_{k}$ is the input dimension.

\textbf{Positional encoding} Transformer \cite{NIPS2017_7181} adds positional encoding (PE) to the input embedding to capture the position information, which is fixed and only related to the absolute position. The $d^{th}$ dimension of the positional encoding in position $pos$ is calculated as:
\begin{align}
    P\!E_{\left ( pos,2d \right )}=\;&\mathrm{sin}\!\left ( pos/10000^{2d/d_{model}} \right )\\
    P\!E_{\left ( pos,2d+1 \right )}=\;&\mathrm{cos}\!\left ( pos/10000^{2d/d_{model}} \right )
\end{align}
where $d_{model}$ is the dimension of input embedding.

\subsection{Simultaneous Machine Translation}

Different from full-sentence MT waiting for the complete sentence, SiMT translates concurrently with the streaming inputs and hence prefix-to-prefix architecture \cite{ma-etal-2019-stacl} is proposed to adapt to the incomplete source, where the target word $y_{i}$ is generated based on a partial source prefix. 

\textbf{Prefix-to-prefix architecture} Let $g(i)$ be a monotonically non-decreasing function of $i$ that denotes the length of received source sentence (i.e., source prefix) when translating the target word $y_{i}$. Given $g(i)$, the probability of generating the target word $y_{i}$ is $p\left ( y_{i}\mid \mathbf{x}_{\leq g(i)},\mathbf{y}_{< i} \right )$, where $\mathbf{x}_{\leq g(i)}$ is first $g(i)$ source words and $\mathbf{y}_{< i}$ is previous target words. Overall, the decoding probability of SiMT is:
\begin{gather}
    p_{\!sim}(\mathbf{y}\mid \mathbf{x})=\prod_{i=1}^{I}p\left ( y_{i}\mid \mathbf{x}_{\leq g(i)},\mathbf{y}_{< i} \right )
\end{gather}

To determine $g(i)$ during translating process, SiMT requires a policy to determine `translating' a target word or `waiting' for the next source word, falling into fixed policy and adaptive policy.

\textbf{Fixed policy} performs `waiting' or `translating' according to pre-defined rules. \emph{Wait-k policy} \cite{ma-etal-2019-stacl} is the most widely used fixed policy, which first waits for $k$ source words and then translates one target word and waits for one source word alternately. Besides, \citet{ma-etal-2019-stacl} also proposed a \emph{test-time wait-k policy}, using a full-sentence model to perform wait-k policy in testing.

\textbf{Adaptive policy} can dynamically adjust `waiting' or `translating' according to the current state. \emph{Monotonic multi-head attention} (\emph{MMA}) \cite{Ma2019a} is the current state-of-the-art adaptive policy, which predicts a Bernoulli action READ/WRITE to decide to wait for the next source word (READ) or translate a target word (WRITE). To train the Bernoulli actions, MMA predicts the writing probability of $y_{i}$ when receiving $x_{j}$, denoted as $\beta _{ij}$, and uses it to approximate the READ/WRITE actions during training \cite{Arivazhagan2019}.

\section{Preliminary Analysis on Position Bias}
\label{sec:positionbias}
In this section, we analyze the influence and cause of position bias in SiMT.
In full-sentence MT, the source sentence is complete, so that each source word participates in the translation of all target words. While in the prefix-to-prefix architecture for SiMT, each target word is forced to align with an increasing source prefix, which directly causes that the source words in the front position participate in the translation of more target words during training and hence are always illusoryly considered more important during testing, resulting in \emph{position bias}. Please refer to Appendix \ref{sec:appendix} for a theoretical analysis of the position bias.

During testing, position bias is reflected in the preference of paying more attention to the source words in front positions. To explore the specific impact of position bias, we select the samples with the same source length (77 sentences) in WMT15 De$\rightarrow$En test set as a bucket, and then calculated the average attention weight obtained by each source position in the bucket. Since the times of each source position being paid attention to may be different in SiMT, the average attention weight is averaged on the times of being attended, so the evaluation is fair for each source position. Specifically, give the attention weight $\alpha _{ij}$ between target word $y_{i}$ and source word $x_{j}$, the average attention weight $\overline{A}_{j}$ at source position $j$ is calculated as:
\begin{gather}
    \overline{A}_{j}=\frac{\sum_{i=1}^{I}\alpha _{ij}}{\sum_{i=1}^{I}\mathbbm{1}_{j\leq g\left ( i \right )}} \label{eq5}
\end{gather}
where $\sum_{i=1}^{I}\alpha _{ij}$ is the sum of attention on the $j^{th}$ source position, and $\sum_{i=1}^{I}\mathbbm{1}_{j\leq g\left ( i \right )}$ counts the times of the $j^{th}$ source position being paid attention to.

\begin{figure}[t]
\centering
\subfigure[ SiMT v.s. Full-sentence MT ]{
\includegraphics[width=2.6in]{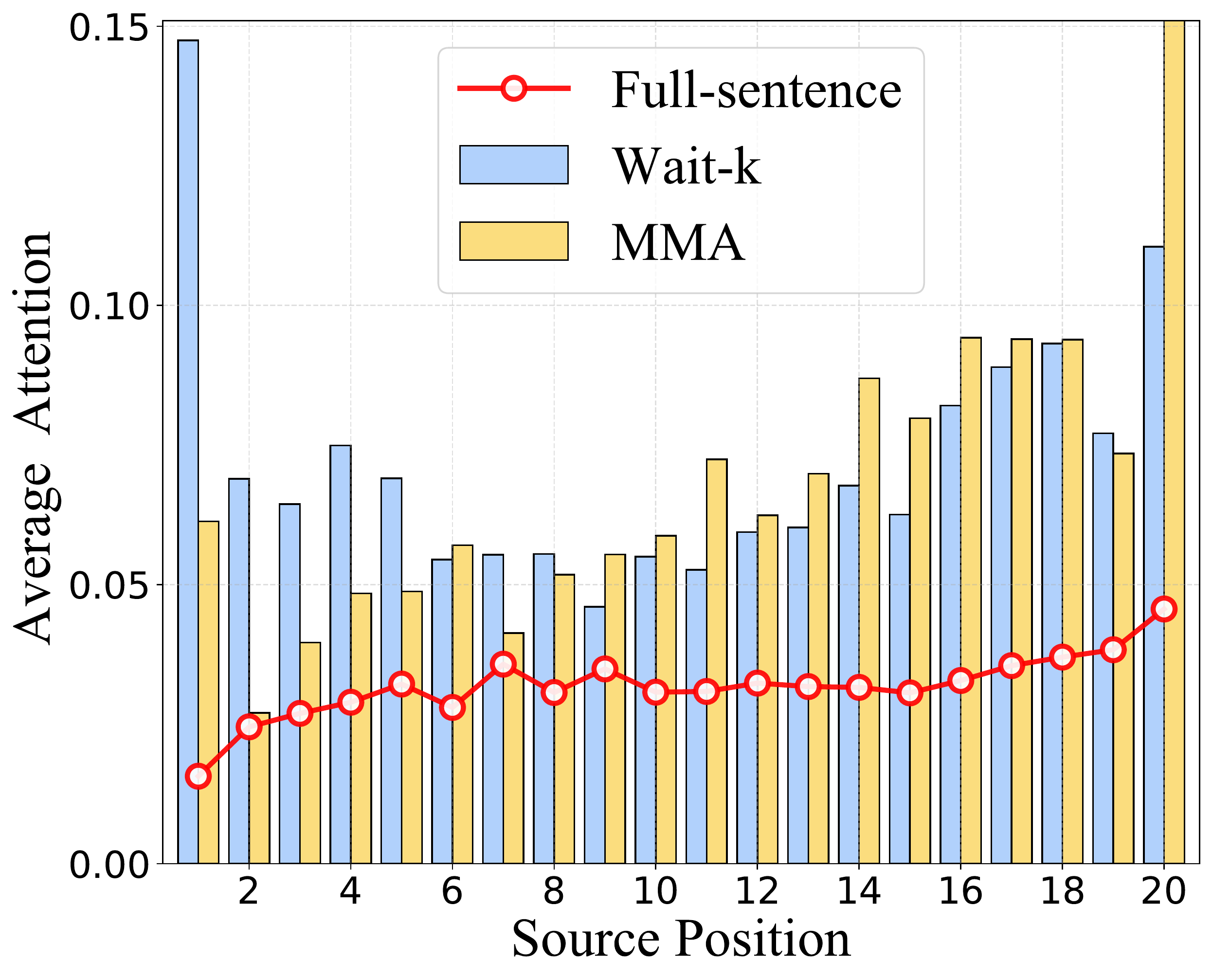}
}
\subfigure[Wait-k v.s. Test-time Wait-k]{
\includegraphics[width=2.6in]{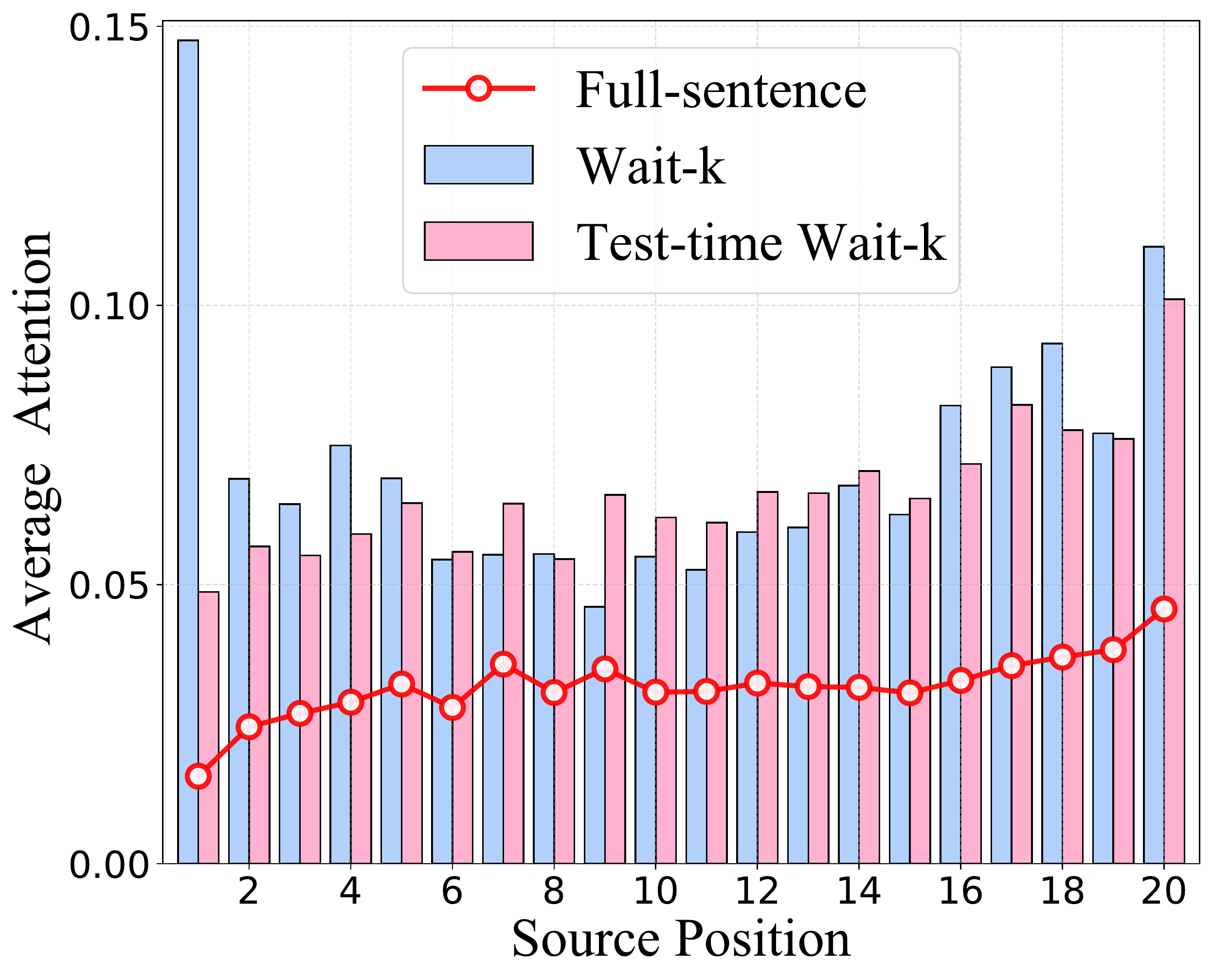}
}

\caption{Average attention $\overline{\mathbf{A}}$ obtained by different source positions on the De$\rightarrow$En task, showing wait-5, test-time wait-k, MMA and full-sentence MT.}
\label{avgattn}
\end{figure}

\begin{figure}[t]
\centering
\subfigure[Full-sentence MT]{
\includegraphics[width=1.47in]{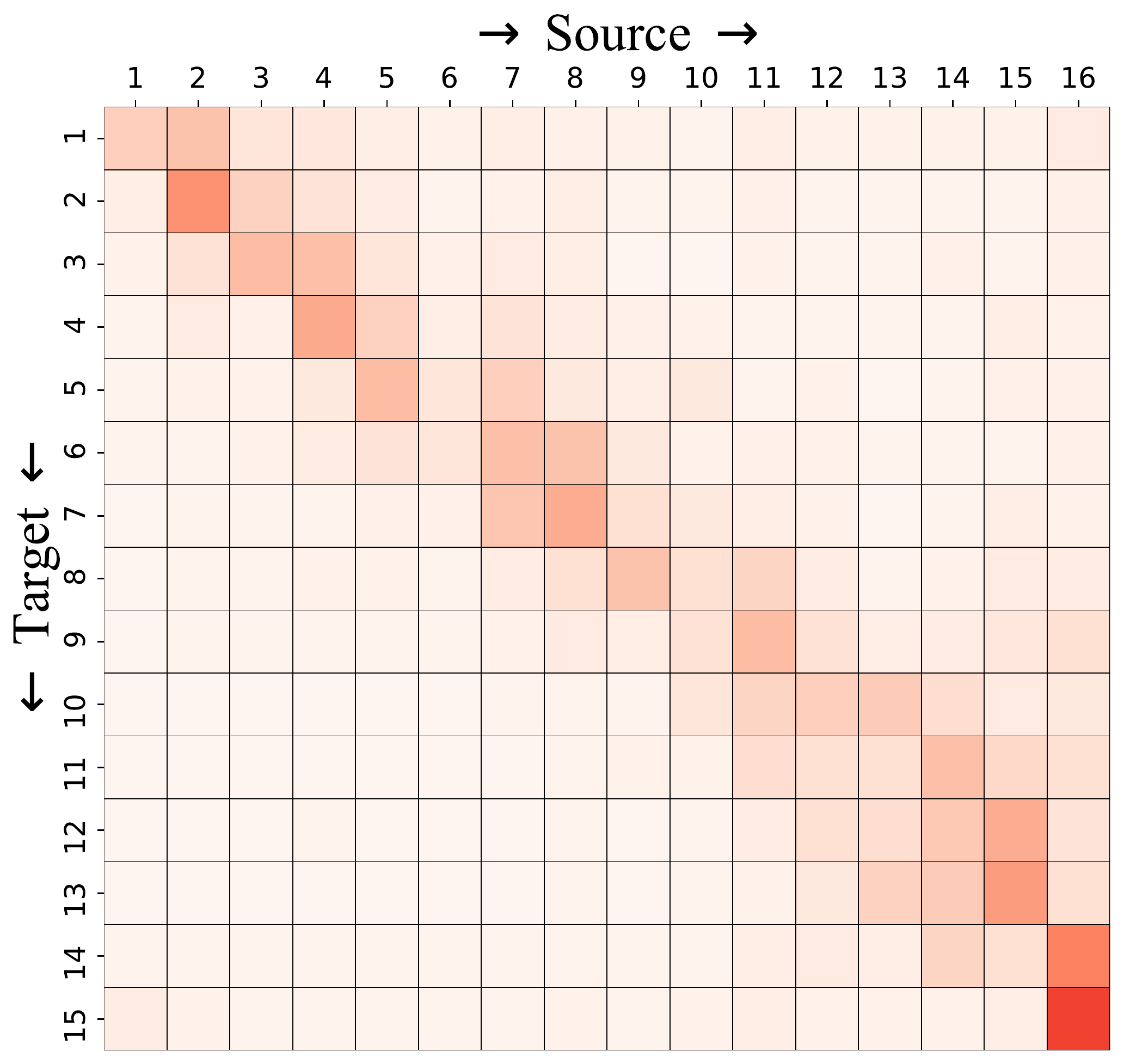}
\label{avgattn1}
}\hspace{-2.5mm}
\subfigure[SiMT with wait-5 policy]{
\includegraphics[width=1.47in]{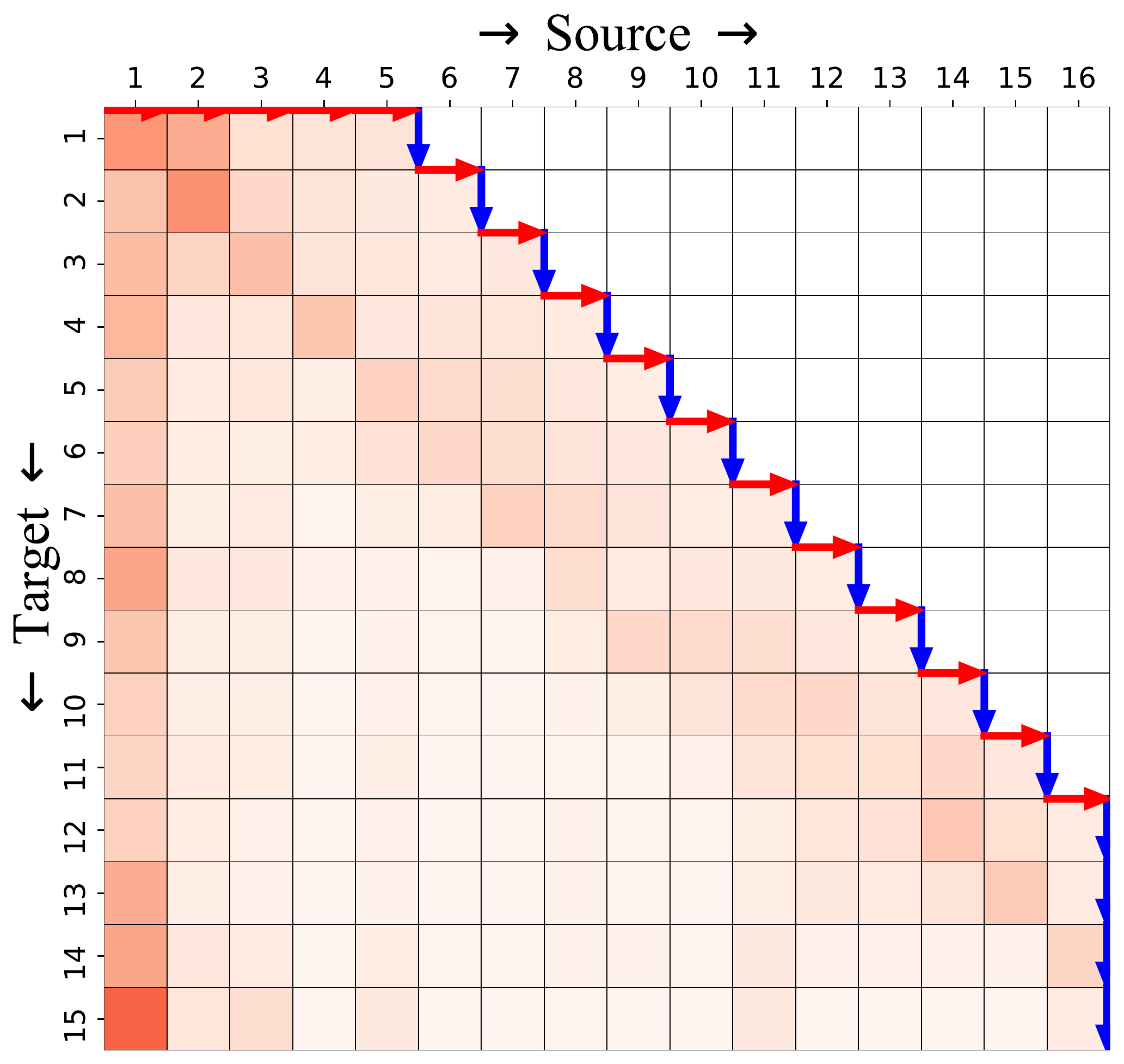}
\label{avgattn2}
}

\caption{Full-sentence MT v.s. SiMT on attention characteristics. We select 20 sentence pairs with the same source and target lengths on De$\rightarrow$En and average their attention matrix to get statistical characteristics. `\textcolor{red}{$\rightarrow$}': wait for a source word, `\textcolor{blue}{$\downarrow$}': translate a target word.}
\label{attn}
\end{figure}

\textbf{What is position bias?} Figure \ref{avgattn1} shows the average attention obtained by different source positions in two representative SiMT methods, compared with full-sentence MT. SiMT has a significant difference from the full-sentence MT on the average attention to the source position. In full-sentence MT, the average attention on each position is similar and the back position gets slightly more attention \cite{voita-etal-2021-analyzing}. However, in both the fix and adaptive policy in SiMT, the front source positions obviously get more attention due to position bias, especially the first source word. Compared with wait-k, MMA alleviates the position bias by dynamically adjusting `waiting' or `translating', but the first source position still abnormally gets more attention. Note that the average attention on the back positions in SiMT is higher since the times they are attended are less (the denominator in Eq.(\ref{eq5}) is smaller).

\textbf{Specific attention characteristics} Furthermore, we compare the characteristics of attention distribution in full-sentence MT and SiMT, shown in Figure \ref{attn}. In SiMT, more attention weights are concentrated on the front source positions \cite{Arivazhagan2019,dualpath}, which is not conducive to translation. First, the biased attention on front positions robs the attention of the aligned source word, resulting in mis-translation error. Second, much overlapping on attention distribution aggravates the duplication translation error, where a human evaluation proposed by \citet{elbayad-etal-2020-online} shows that duplication error in SiMT is 500\% of full-sentence MT. Besides, in some cases, even if the aligned source words have not been received, the prefix-to-prefix architecture still forces the target word to align with the irrelevant source prefix, resulting in the confusion on attention \cite{chen-etal-2021-improving-simultaneous}.

\begin{figure}[t]
\centering
\subfigure[Divided based on position bias degree in wait-k.]{
\includegraphics[width=2.6in]{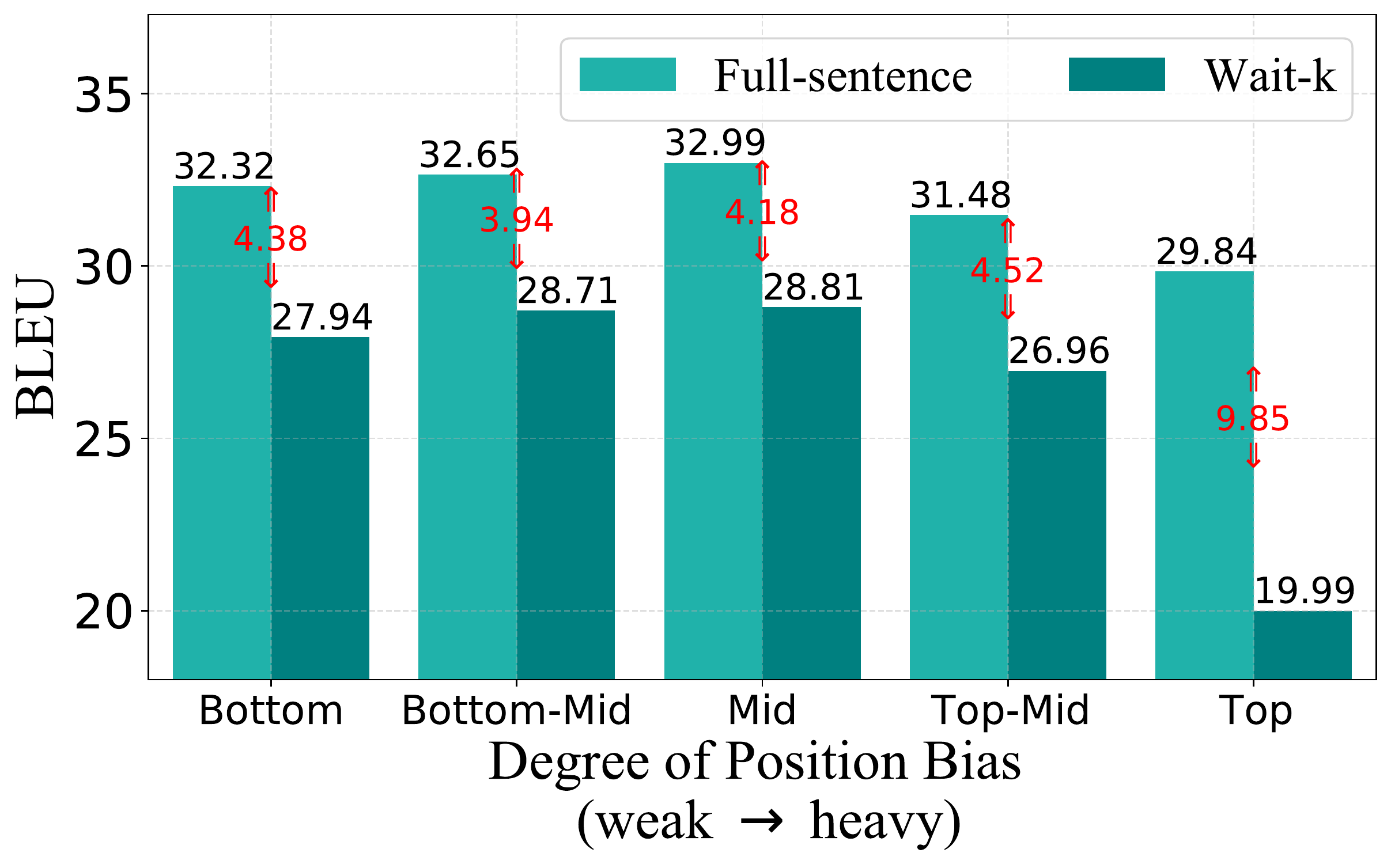}
}
\subfigure[Divided based on position bias degree in MMA.]{
\includegraphics[width=2.6in]{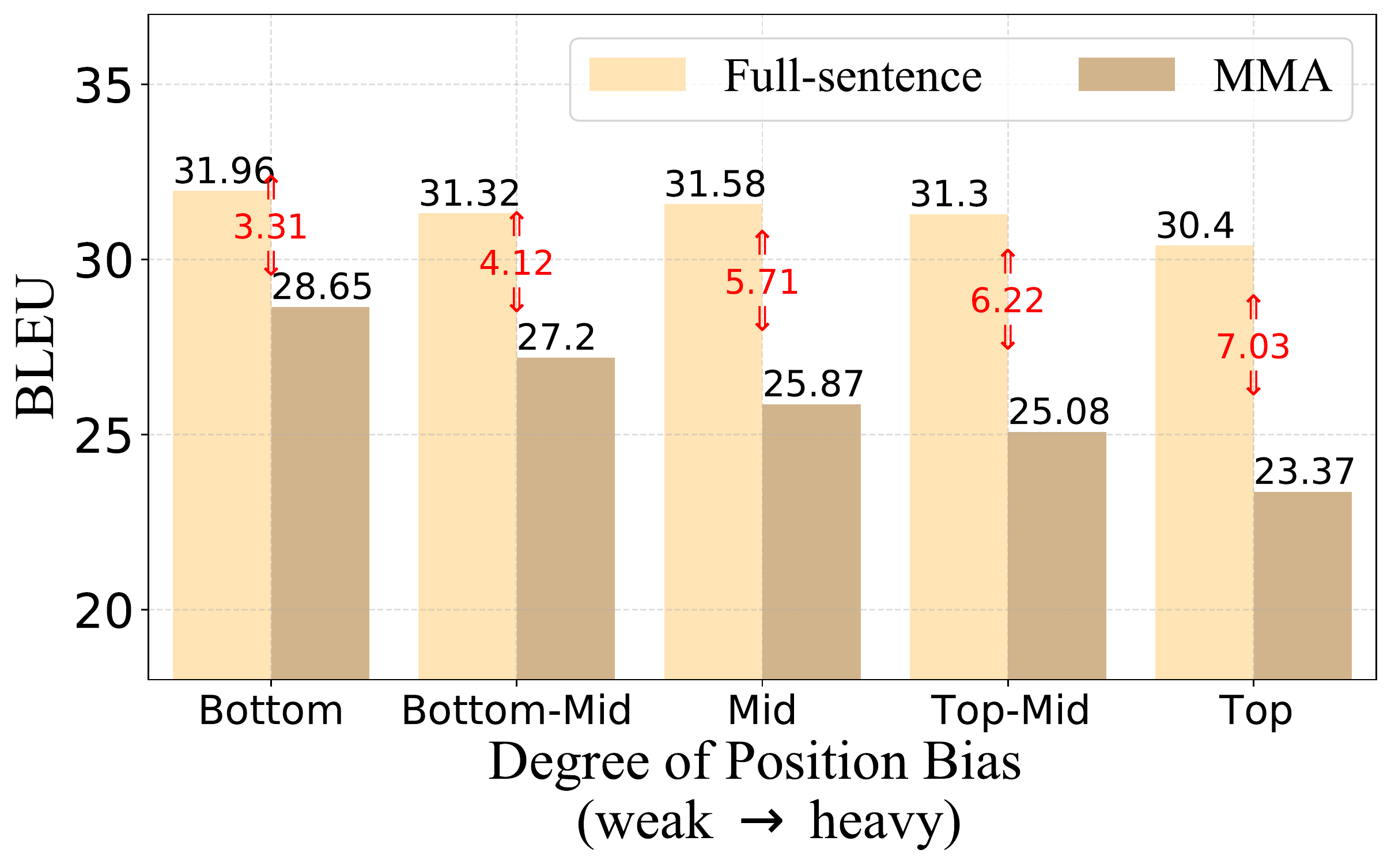}
}

\caption{Performance with degree of position bias. }
\label{bleu_bias}
\end{figure}

\textbf{Does position bias affect SiMT performance?} To analyze whether the position bias in SiMT results in poor translation quality, we use the ratio of the average attention on the first source position to all positions ($\overline{A}_{1}/\sum_{j}\overline{A}_{j}$) to reflect the degree of position bias, and accordingly divide WMT15 De$\rightarrow$En test set into 5 parts evenly. We report the translation quality of these 5 parts in Figure \ref{bleu_bias}, where the position bias is heavier from `Bottom' to `Top'. The translation quality of both wait-k and MMA  significantly decrease as the position bias becomes heavy, while full-sentence MT remained high-quality translation on these parts. More importantly, as the position bias intensifies, the performance gap between SiMT and full-sentence MT is amplified, where wait-k and MMA are 9.85 BLEU and 7.03 BLEU lower than full-sentence MT respectively on the `Top' set. Therefore, the position bias is an important cause of the performance gap between SiMT and full-sentence MT.

\textbf{What is the position bias caused by?} To verify that the preference for front source positions is caused by the structural gap between SiMT and full-sentence MT rather than streaming inputs during testing, we compare the average attention of wait-k and `test-time wait-k' in Figure \ref{avgattn2}, where `test-time wait-k' is trained with full-sentence structure and tested with wait-k policy. After replacing the prefix-to-prefix architecture with the seq-to-seq architecture during training, the position bias in the `test-time wait-k' is significantly weakened, which shows that prefix-to-prefix training is the main cause of position bias. However, directly training with full-sentence structure leaks many future source words, where the obvious training-testing mismatch results in inferior translation quality of `test-time wait-k' \cite{ma-etal-2019-stacl}.

In practice, prefix-to-prefix architecture forces the target word to assign attention to the prefix even if its corresponding source word has not been read in, which will undoubtedly cause the attention to become chaotic and tend to be distributed to the front position. This also explains why the position bias is more serious in the fixed policy, since the read/write cannot be adjusted, in more cases the prefix does not contain the corresponding source word but is forced to pay attention to. Besides, prefix-to-prefix architecture increases the frequency of front source positions during training, and previous works \cite{1549828,luong-etal-2015-addressing,gu-etal-2020-token} show that NMT models have a tendency towards over-fitting on high-frequency words, resulting in the position bias.

\section{The Proposed Method}

\begin{figure}[t]
\centering
\includegraphics[width=2.9in]{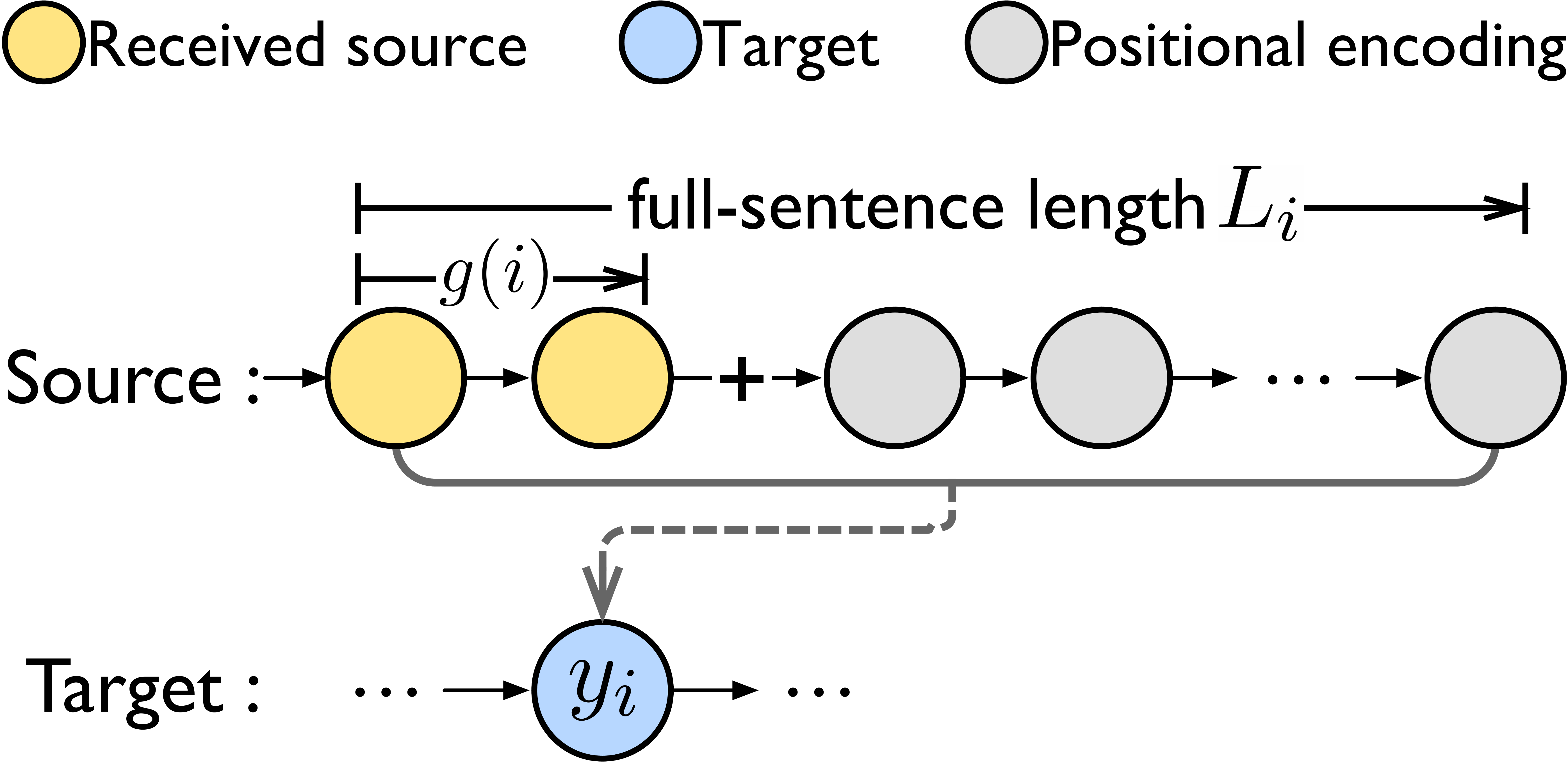}
\caption{Length-aware framework for SiMT, which first predicts full-sentence length $L_{i}$ and fills the future source position with positional encoding. }
\label{laf}
\end{figure}

Based on the preliminary analyses on position bias, we hope that in SiMT, the target words can also align to reasonable source positions as them in full-sentence MT, including the future positions even though the words on these positions have not yet been received. Along this line, we develop a \emph{Length-Aware Framework} (\emph{LAF}) to turn the streaming inputs into pseudo full-sentence and thereby allow the target words to align with the full-sentence positions rather than a prefix, as shown in Figure \ref{laf}. The details are introduced following.

\subsection{Length-Aware Framework}
\label{sec:LAF}

\textbf{Length prediction} 
To turn the incomplete source into pseudo full-sentence, LAF first predicts the full-sentence length. At step $i$, based on the current received source sentence $\mathbf{x}_{\leq g\left ( i \right )}$, LAF predicts the full-sentence length $L_{i}$ through a classification task. Note that the predicted length dynamically changes with the increase of received source words.

Formally, the probability of full-sentence length $L_{i}$ is predicted through a multi-layer perceptron (MLP) based on the mean of hidden states of the currently received source words: 
\begin{gather}
    \!p_{l}\!\left ( L_{i}\!\mid\! \mathbf{x}_{\leq g\left ( i \right )} \right )\!\!=\!\mathrm{softmax}\!\left ( \! \mathbf{W}\mathrm{tanh}\!\left (\! \mathbf{V}\overline{h}_{\leq g(i)} \right )\!\right )
\end{gather}
where $\overline{h}_{\leq g(i)}\!=\!\frac{1}{g(i)}\sum_{j=1}^{g(i)}\!h_{j}$ is the the mean of hidden states of the currently received source words. $\mathbf{V} \!\!\in \!\!\mathbb{R}^{d_{model}\!\times \!d_{model}}$ and $\mathbf{W}\!\! \in \!\!\mathbb{R}^{N_{max}\!\times\! d_{model}}$ is the parameters of MLP, where $N_{max}$ is the max length of the source sentence in the corpus. Note that $\mathrm{softmax}(\cdot)$ is normalized on all possible length values. In testing, the value with the highest probability is selected as the full-sentence length.

If source sentence is already complete (receiving $\left \langle \mathrm{eos} \right \rangle$) or the predicted length $L_{i}$ is not larger than the received source length ($L_{i}\leq g(i)$), we use the current length $g(i)$ as the full-sentence length.

\textbf{Pseudo full-sentence}
Given the predicted full-sentence length, we fill the future source position $\left (g\left ( i \right ),L_{i} \right ]$ with positional encoding to construct the pseudo full-sentence. Formally, given the hidden states of received source word $h_{\leq g(i)}$ and the predicted full-sentence length $L_{i}$, we fill the future position with positional encoding to get the pseudo full-sentence hidden states $\widetilde{\mathbf{h}}^{(i)}$ at step $i$:
\begin{gather}
    \widetilde{\mathbf{h}}^{(i)}=\left ( h_{1},\cdots ,h_{g(i)},P\!E_{g(i)+1},\cdots,P\!E_{L_{i}} \right )
\end{gather}

Then, the target word $y_{i}$ is generated based on the pseudo full-sentence hidden states $\widetilde{\mathbf{h}}^{(i)}$ and previous target word $y_{<i}$, and hence the cross-attention $\alpha _{ij}$ in Eq.(\ref{eq1}) is rewritten as:
\begin{gather}
    \alpha _{ij}=\mathrm{softmax}\left ( \frac{s_{i}W^{Q}\left (\widetilde{h}^{(i)}_{j}W^{K}  \right )^{\top} }{\sqrt{d_{k}}} \right )
\end{gather}
Overall, the decoding probability of the length-aware framework is:
\begin{gather}
\begin{aligned}
    p_{laf}(\mathbf{y}\mid \mathbf{x})=\prod_{i=1}^{I} \, p_{l}&\left ( L_{i}\mid  \mathbf{x}_{\leq g(i)}\right )  \times \\ 
    p&\left ( y_{i}\mid \mathbf{x}_{\leq g(i)},\mathbf{y}_{< i},L_{i} \right ) \label{eq8}
\end{aligned}
\end{gather}

\subsection{Training Objective}
\label{sec:training}
The length-aware framework consists of a length prediction module and a translation module. For the length prediction module, we take the complete source length $J$ as the ground-truth length label and train the model with cross-entropy loss:
\begin{gather}
    \mathcal{L}_{len}=-\sum_{i=1}^{I}\mathrm{log}\;p_{l}\left ( J\mid \mathbf{x}_{\leq g\left ( i \right )} \right )
\end{gather}

For the translation module, we complement the source prefix to the ground-truth source length $J$ with positional encoding and train the translation module by minimizing the cross-entropy loss:
\begin{gather}
    \mathcal{L}_{ce}=-\sum_{i=1}^{I}\mathrm{log}\;p\left ( y_{i}^{\star }\mid  \mathbf{x}_{\leq g(i)},\mathbf{y}^{\star }_{< i}, J  \right )
\end{gather}
where $\mathbf{y}^{\star }$ is the ground-truth target sentence. During testing, we apply the predicted full-sentence length to complement the source prefix. We will compare the performance of training with ground-truth or predicted full-sentence length in Sec.\ref{sec:ablation}. 

Finally, the total loss of LAF is calculated as:
\begin{gather}
    \mathcal{L}_{laf}=\mathcal{L}_{ce}+\mathcal{L}_{len}
\end{gather}

\subsection{Integrated into SiMT Policy}
\label{sec:inte}
The length-aware framework can be integrated into most existing SiMT methods. We take wait-k and MMA as representatives to introduce the slight difference when integrated to fix and adaptive policy respectively. LAF predicts the full-sentence length based on the currently received source words $\mathbf{x}_{\leq g\left ( i \right )}$, so the key is to calculate $g\left ( i \right )$, which may be different in fix and adaptive policy.

\textbf{Fixed policy} Since wait-k is a pre-defined fixed policy, $g_{wait-k}\left ( i \right )$ in wait-k during both training and testing is invariably calculated as:
\begin{gather}
    g_{wait-k}\left ( i \right )=\min \left \{ k+i-1, \left | \mathbf{x} \right | \right \}
\end{gather}

\textbf{Adaptive policy} Since MMA can dynamically predict READ/WRITE actions, the calculation of $g\left ( i \right )$ during training and testing is different. During testing, we take the number of source words received by the model when starting to translate $y_{i}$ as $g\left ( i \right )$. During training, MMA does not have explicit READ/WRITE actions, but predicts the writing probability $\beta _{ij}$, where $\beta _{ij}$ represents the probability of translating $y_{i}$ after receiving source word $x_{j}$. Therefore, we select the position of $x_{j}$ with the highest writing probability as $g_{mma}\left ( i \right )$:
\begin{gather}
    g_{mma}\left ( i \right )=\underset{j}{\mathrm{argmax}}\;\beta _{ij}
\end{gather}

\section{Related Work}
The main architectures of SiMT model are divided into two categories: seq-to-seq architecture and prefix-to-prefix architecture.

The early SiMT methods always used a full-sentence MT model trained by seq-to-seq architecture to translate each segment divided by the SiMT policy \cite{bangalore-etal-2012-real,Cho2016,siahbani-etal-2018-simultaneous}. \citet {gu-etal-2017-learning} used reinforcement learning to train an agent to decide whether to start translating. \citet {Alinejad2019} added a predict operation based on \citet {gu-etal-2017-learning}. \citet{zhang-etal-2020-learning-adaptive} proposed an adaptive segmentation policy based on meaning units. However, the mismatch between training and testing usually leads to inferior translation quality.

The recent SiMT methods, including fix and adaptive policies, mainly used prefix-to-prefix architecture. For the fixed policy, \citet {ma-etal-2019-stacl} proposed a wait-k policy, which always generates target token $k$ words behind the source token. \citet{zhang-feng-2021-icts} proposed a char-level wait-k policy. \citet{zhang-feng-2021-universal} proposed a universal SiMT with the mixture-of-experts wait-k policy. For the adaptive policy, \citet{Zheng2019b} trained an agent with the golden read/write action sequence. \citet {Zheng2019a} added a ``delay'' token and introduced limited dynamic prediction. \citet {Arivazhagan2019} proposed MILk, using a Bernoulli variable to determine whether to write. \citet {Ma2019a} proposed MMA to implement MILK on the Transformer. \citet{wilken-etal-2020-neural} and \citet{gma} proposed alignment-based SiMT policy. \citet{liu-etal-2021-cross} proposed cross-attention augmented transducer for SiMT. \citet{future-guided} and \citet{alinejad-etal-2021-translation} introduced a full-sentence model to guide SiMT policy. \citet{miao-etal-2021-generative} proposed a generative SiMT policy.

Although the prefix-to-prefix architecture simulates the streaming inputs, it brings the position bias described in Sec.\ref{sec:positionbias}. Therefore, we proposed a length-aware framework to reduce the position bias and meanwhile fulfill the streaming inputs.

\section{Experiments}

\subsection{Datasets}

We evaluate LAF on the following datasets.

\begin{figure*}[t]
\centering
\subfigure[En$\rightarrow$Vi, Transformer-Small]{
\includegraphics[width=1.9in]{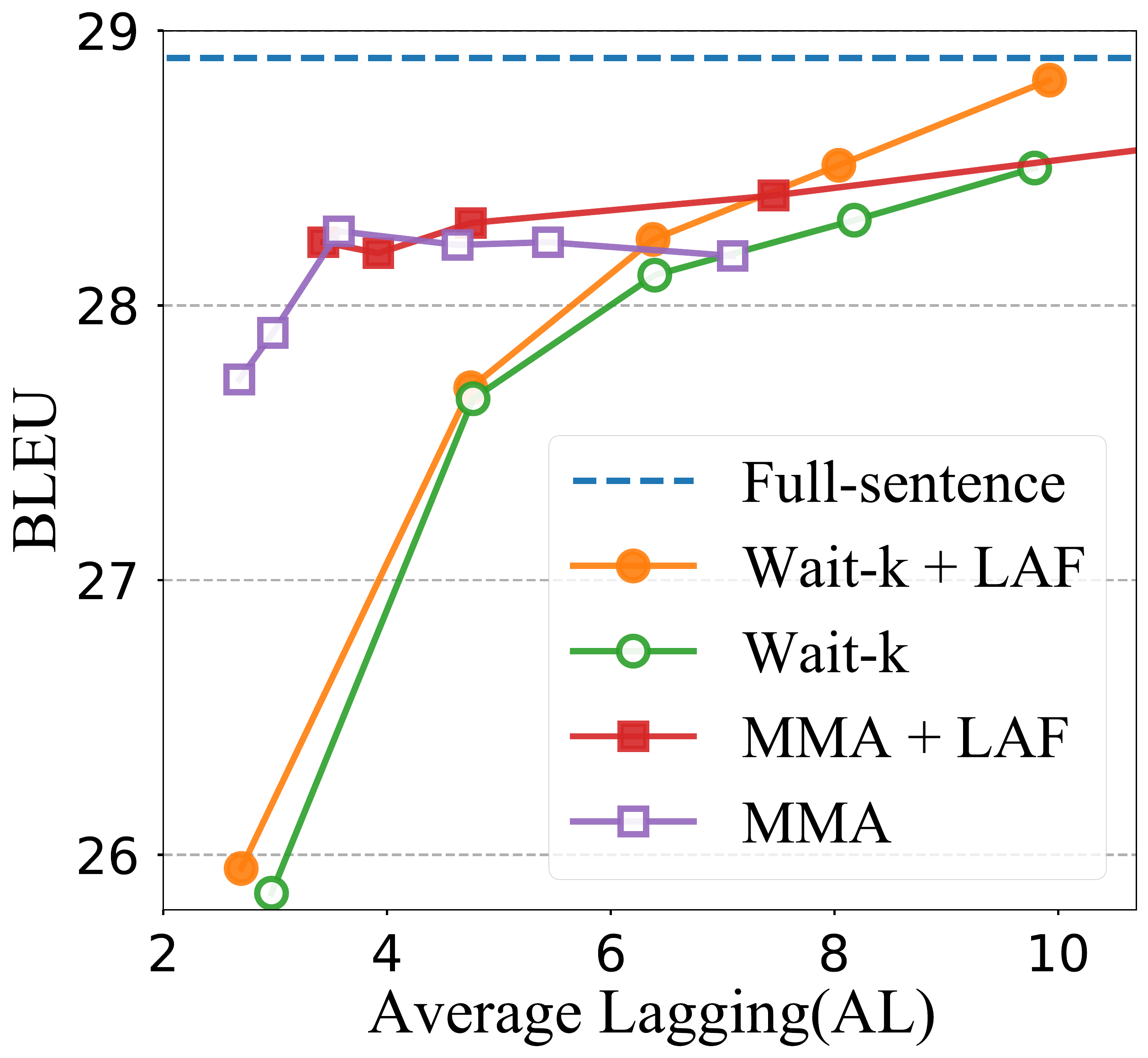}
}
\subfigure[De$\rightarrow$En, Transformer-Base]{
\includegraphics[width=1.9in]{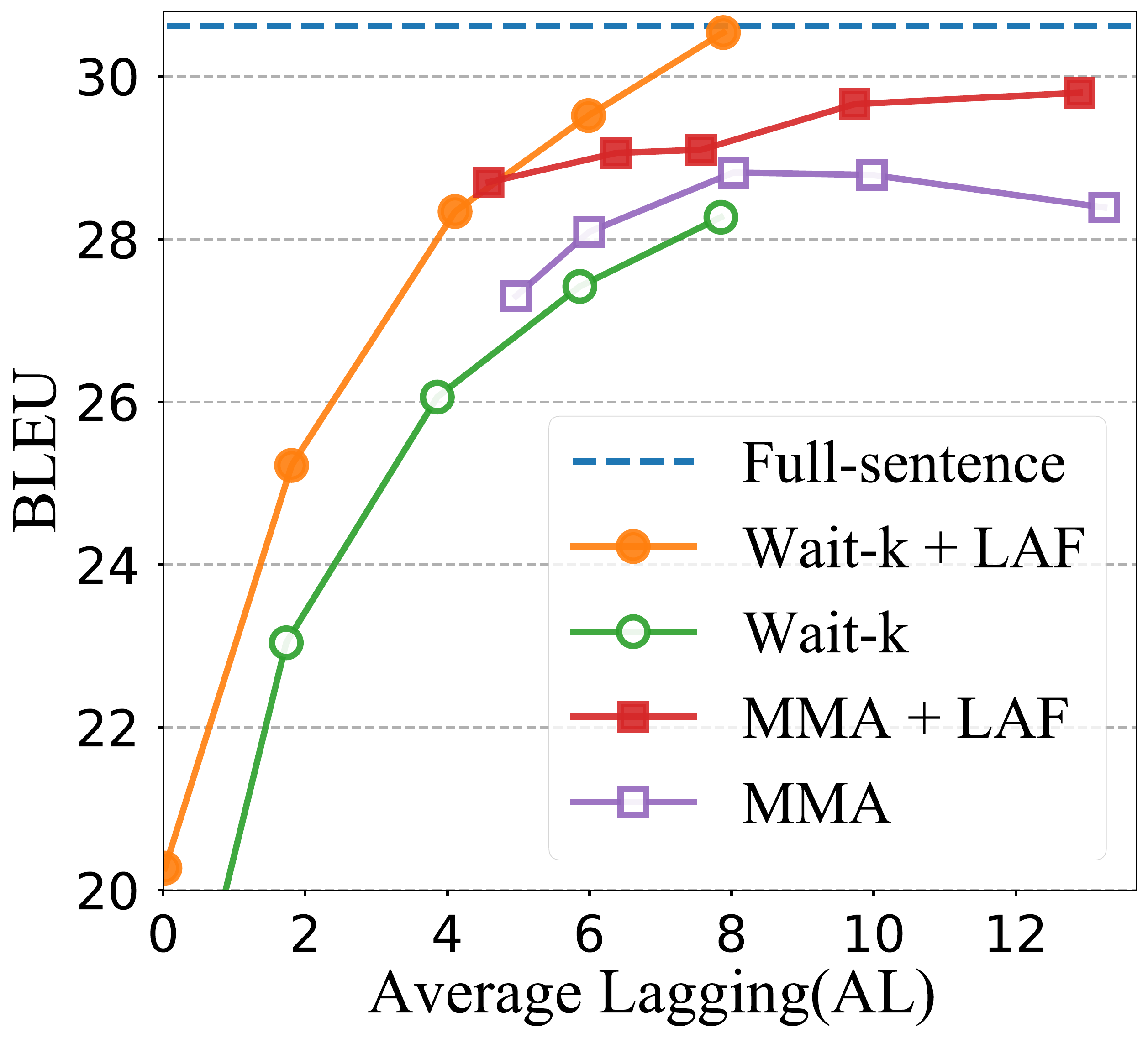}
}
\subfigure[De$\rightarrow$En, Transformer-Big]{
\includegraphics[width=1.9in]{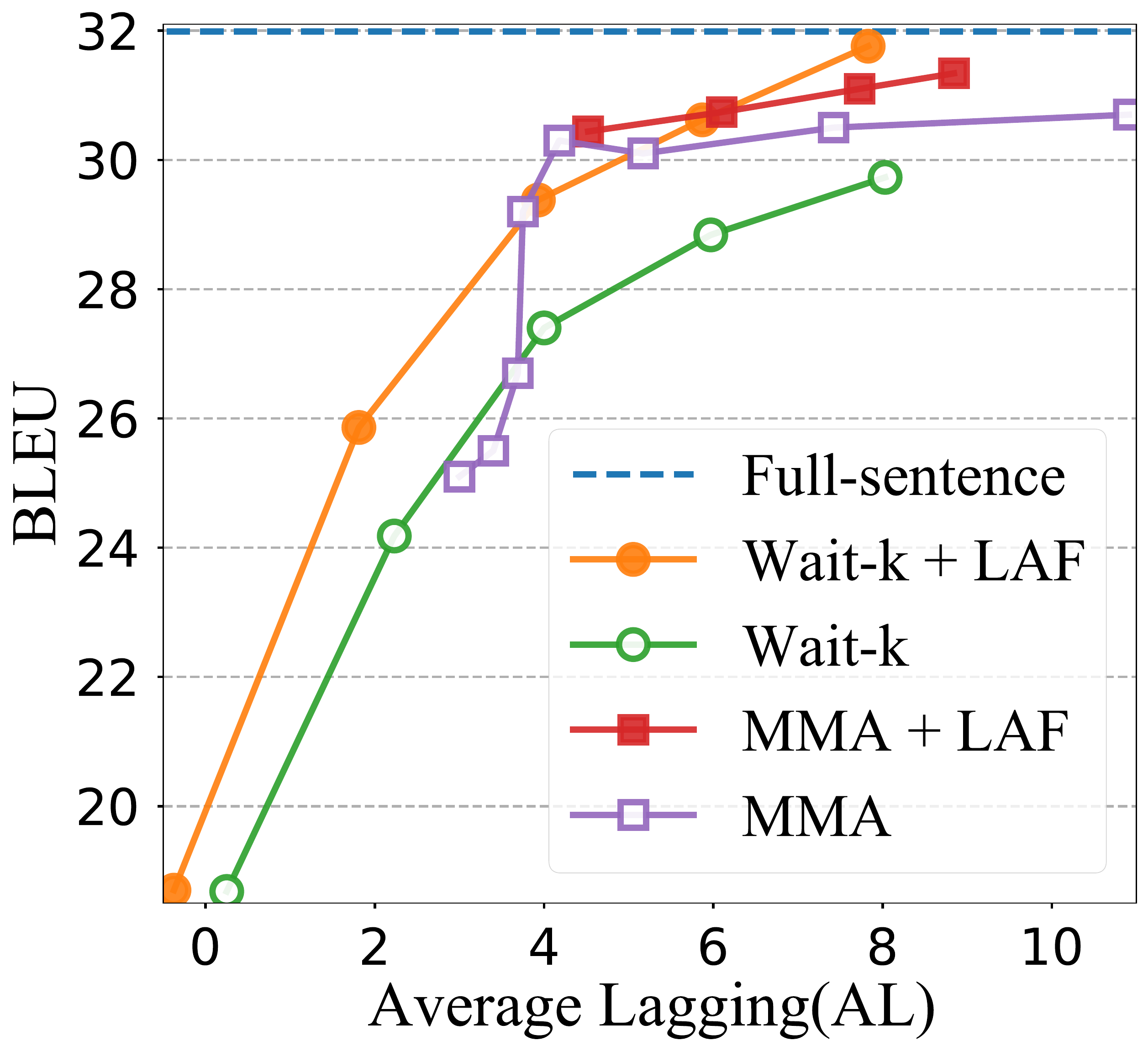}
}
\caption{Translation quality (BLEU) against latency (AL) on the En$\rightarrow$Vi(Small), De$\rightarrow$En(Base) and De$\rightarrow$En(Big).}
\label{main}
\end{figure*}

\textbf{IWSLT15\footnote{\url{nlp.stanford.edu/projects/nmt/}} English$\rightarrow $Vietnamese (En$\rightarrow$Vi)} (133K pairs) \cite{iwslt2015} We use TED tst2012 as validation set (1553 pairs) and TED tst2013 as test set (1268 pairs). Following the previous setting \cite{LinearTime,Ma2019a}, we replace tokens that the frequency less than 5 by $\left \langle unk \right \rangle$, and the vocabulary sizes are 17K and 7.7K for English and Vietnamese respectively.

\textbf{WMT15\footnote{\url{www.statmt.org/wmt15/translation-task}} German$\rightarrow $English (De$\rightarrow$En)} (4.5M pairs) Following \citet{ma-etal-2019-stacl}, \citet{Arivazhagan2019} and \citet{Ma2019a}, we use newstest2013 as validation set (3000 pairs) and newstest2015 as test set (2169 pairs). BPE \cite{sennrich-etal-2016-neural} was applied with 32K merge operations and the vocabulary is shared across languages.

\subsection{Systems Setting}
We conduct experiments on following systems.

{\bf Full-sentence} Full-sentence MT with standard Transformer \cite{NIPS2017_7181}.

{\bf Wait-k} Wait-k policy proposed by \citet{ma-etal-2019-stacl}, the most widely used fixed policy, which first waits for $k$ source words and then translates a target word and waits for a source word alternately.

{\bf {MMA}\footnote{\url{github.com/pytorch/fairseq/tree/master/examples/simultaneous_translation}}} Monotonic multi-head attention (MMA) proposed by \cite{Ma2019a}, the SOTA adaptive policy. At each step, MMA predicts a Bernoulli variable to decide whether to start translating.

{\bf * + LAF} Applying proposed length-aware framework on Wait-k or MMA.

The implementation of all systems are adapted from Fairseq Library \cite{ott-etal-2019-fairseq} based on Transformer \cite{NIPS2017_7181} with the same setting in \citet {Ma2019a}. For En$\rightarrow$Vi, we apply Transformer-small (4 heads). For De$\rightarrow$En, we apply Transformer-Base (8 heads) and Transformer-Big (16 heads). We evaluate these systems with BLEU \cite{papineni-etal-2002-bleu} for translation quality and Average Lagging (AL) \cite{ma-etal-2019-stacl} for latency. AL is calculated based on $g\left ( t \right )$:
\begin{gather}
    \mathrm{AL}=\frac{1}{\tau }\sum_{t=1}^{\tau}g\left ( t \right )-\frac{t-1}{\left | \mathbf{y} \right |/\left | \mathbf{x} \right |}
\end{gather}
where $\tau =\underset{t}{\mathrm{argmax}}\left ( g\left ( t \right )= J\right )$. $I$ and $J$ are target and source length respectively.

\subsection{Main Results}

Figure \ref{main} shows the performance improvement that LAF brings to Wait-k and MMA, where our method achieves higher translation quality under all latency. LAF has a more significant improvement on the fixed policy Wait-k, improving about 0.28 BLEU on En$\rightarrow $Vi, 1.94 BLEU on De$\rightarrow $En(Base), 1.50 BLEU on De$\rightarrow $En(Big), which is because the position bias in original wait-k is more serious. Compared with the SOTA adaptive policy MMA, our method also performs better and is much closer to full-sentence MT performance.

\section{Analysis}

We conduct extensive analyses to understand the specific improvements of our method. Unless otherwise specified, all the results are reported on De$\rightarrow$En(Base) and tested with wait-5 (AL=4.10) and MMA (AL=4.57) under similar latency.

\subsection{Ablation Study}
\label{sec:ablation}

\begin{table}[]
\centering
\begin{tabular}{l|cc|cc} \hline
           & \textbf{Train} & \textbf{Test} & \textbf{AL}   & \textbf{BLEU}  \\ \hline
LAF        & GT    & Pred & 4.11 & 28.34 \\ \hdashline
Pred LAF   & Pred  & Pred & 4.07 & 28.21 \\
Oracle LAF & GT    & GT   & 3.93 & 28.37 \\ \hline
\end{tabular}
\caption{An ablation study of using predicted full-sentence length (Pred) or ground-truth source length (GT) in training and testing respectively, where the results are based on the wait-5 policy.}
\label{ablation}
\end{table}

We use ground-truth full-sentence length to train the translation module, and use the predicted full-sentence length in testing. We conduct the ablation study of using predicted full-sentence length (Pred) or ground-truth length (GT) for translation in training and testing respectively, reported in Table \ref{ablation}. 

LAF has a better performance than `Pred LAF', indicating that using ground-truth length during training is more helpful for learning translation. Compared with `Oracle LAF' that uses ground-truth full-sentence length in testing, LAF achieves comparable performance, which shows that the length prediction module in LAF performs well.

\subsection{Accuracy of Predicted Length}

\begin{figure}[t]
\centering
\subfigure[Acc. with latency]{
\includegraphics[width=1.47in]{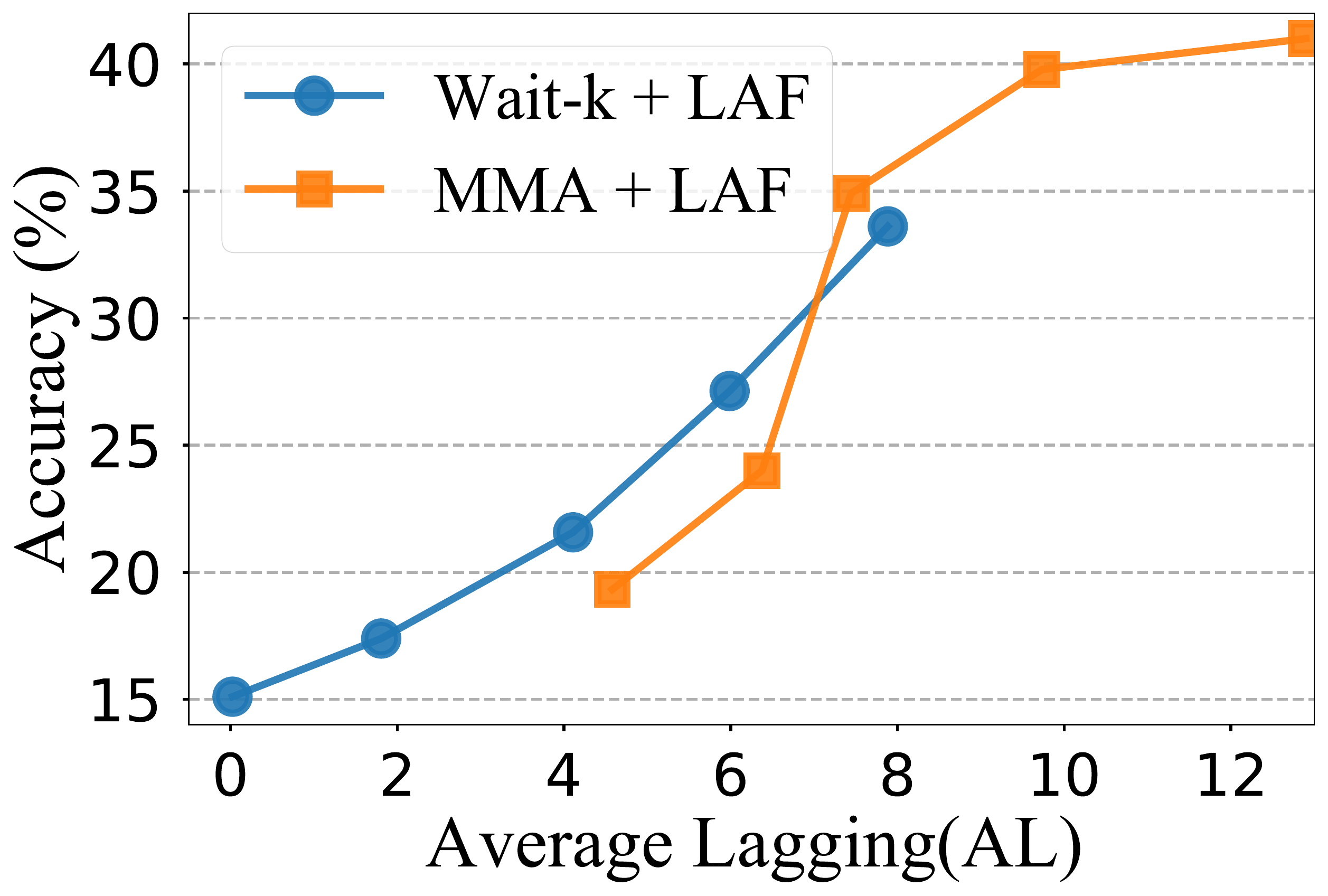}
\label{lenacc1}
}\hspace{-2.5mm}
\subfigure[Acc. with \#received words]{
\includegraphics[width=1.47in]{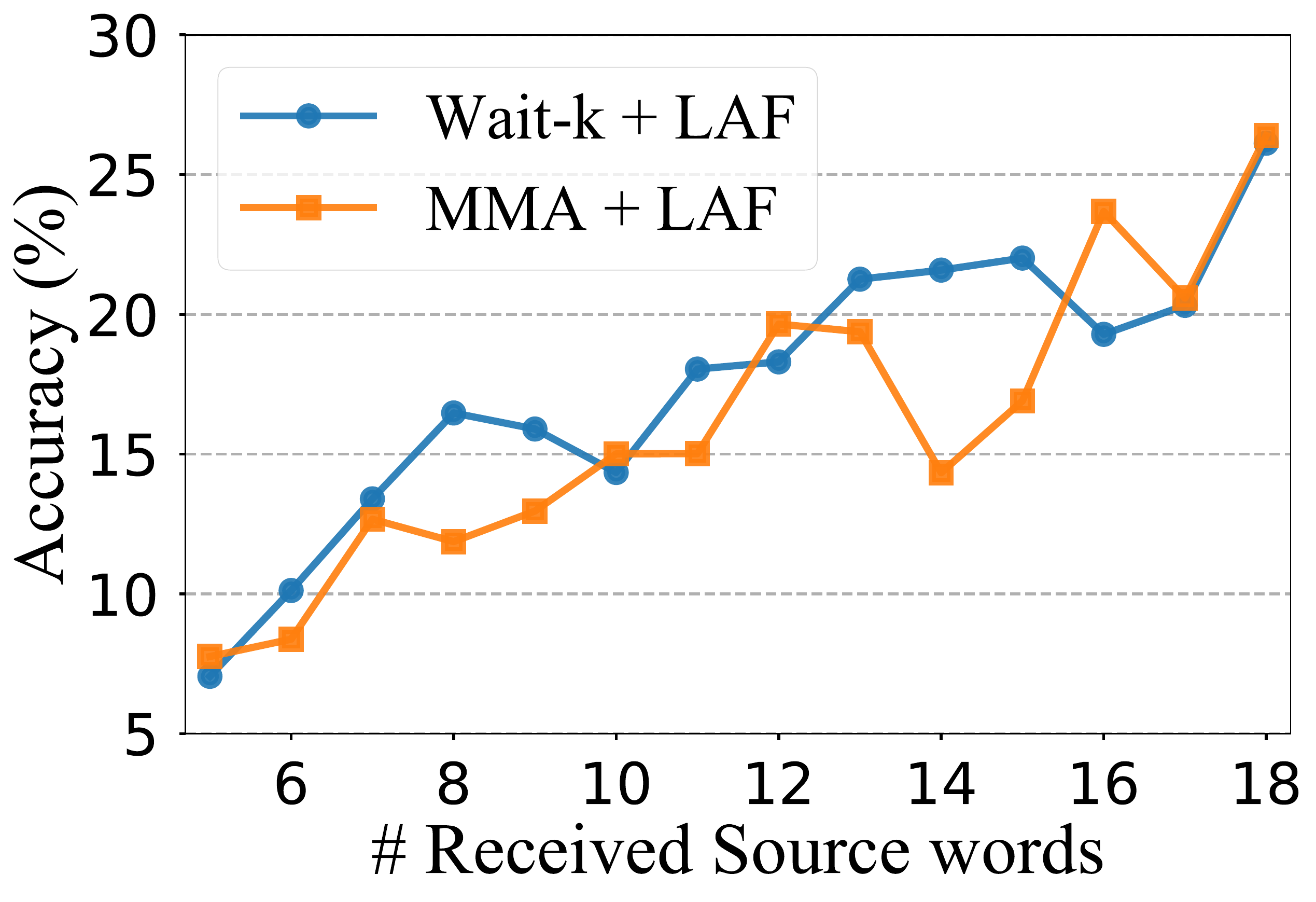}
\label{lenacc2}
}

\caption{Accuracy of predicted length in LAF. (a) Prediction accuracy under different latency. (b) The prediction accuracy with the increasing number of received source words, showing wait-5 and MMA (AL=4.57).}
\label{lenacc}
\end{figure}

Figure \ref{lenacc1} shows the prediction accuracy of the full-sentence length in LAF, indicating that our method achieves good prediction performance. As the latency increases, the prediction accuracy of both `Wait-k+LAF' and `MMA+LAF' gradually increases. Specifically, `Wait-k+LAF' predicts more accurately at low latency, which shows that the regular form of fixed policy is more conducive to LAF learning the full-sentence length. Besides, in Figure \ref{lenacc2}, with the continuous increase of received source words, the prediction accuracy of the full-sentence length gradually improves, which is in line with our expectations.

\subsection{Reduction of Position Bias}

\begin{figure}[t]
\centering
\subfigure[Wait-k+LAF v.s. Wait-k]{
\includegraphics[width=1.47in]{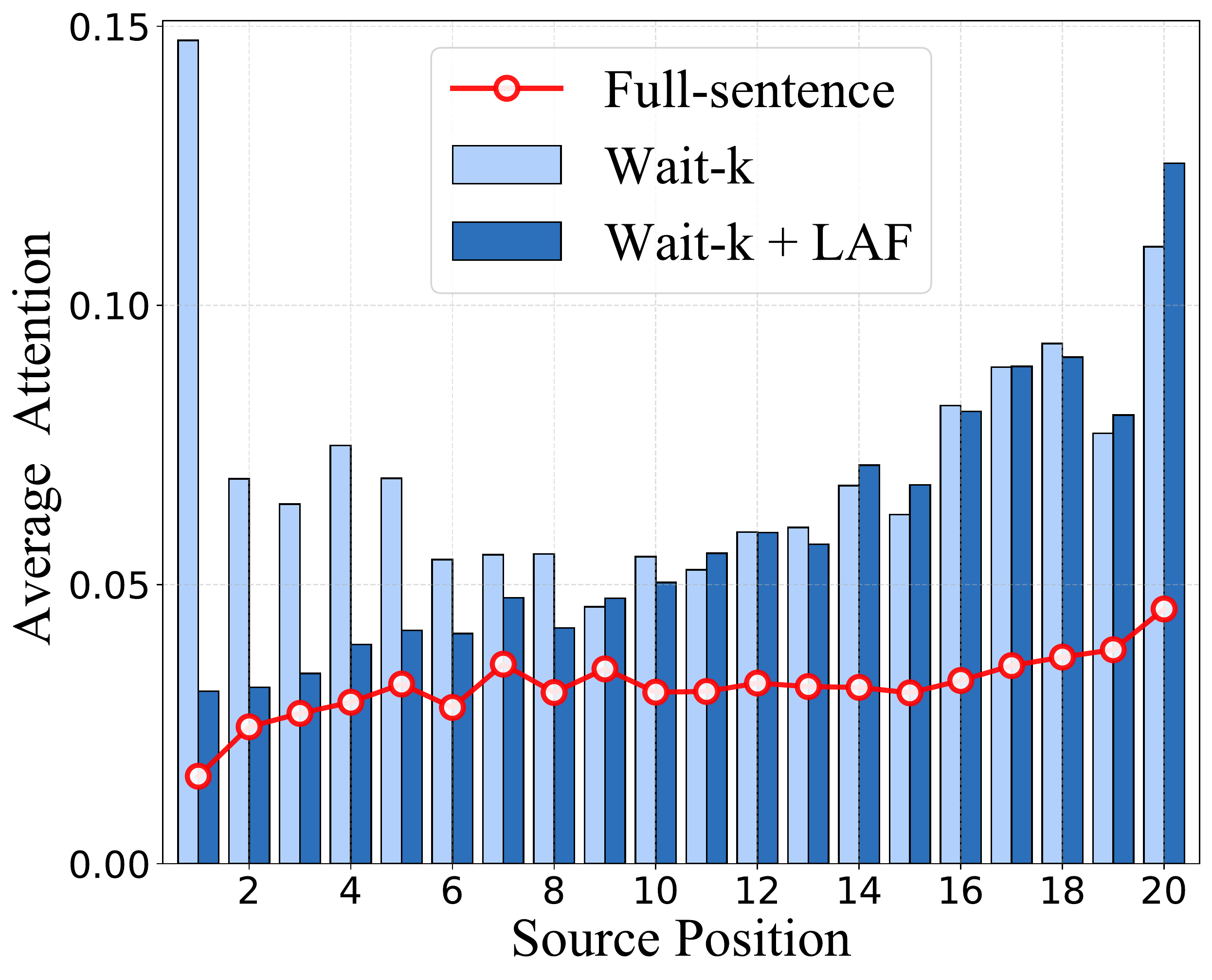}
\label{LAFpb1}
}\hspace{-2.5mm}
\subfigure[MMA+LAF v.s. MMA]{
\includegraphics[width=1.47in]{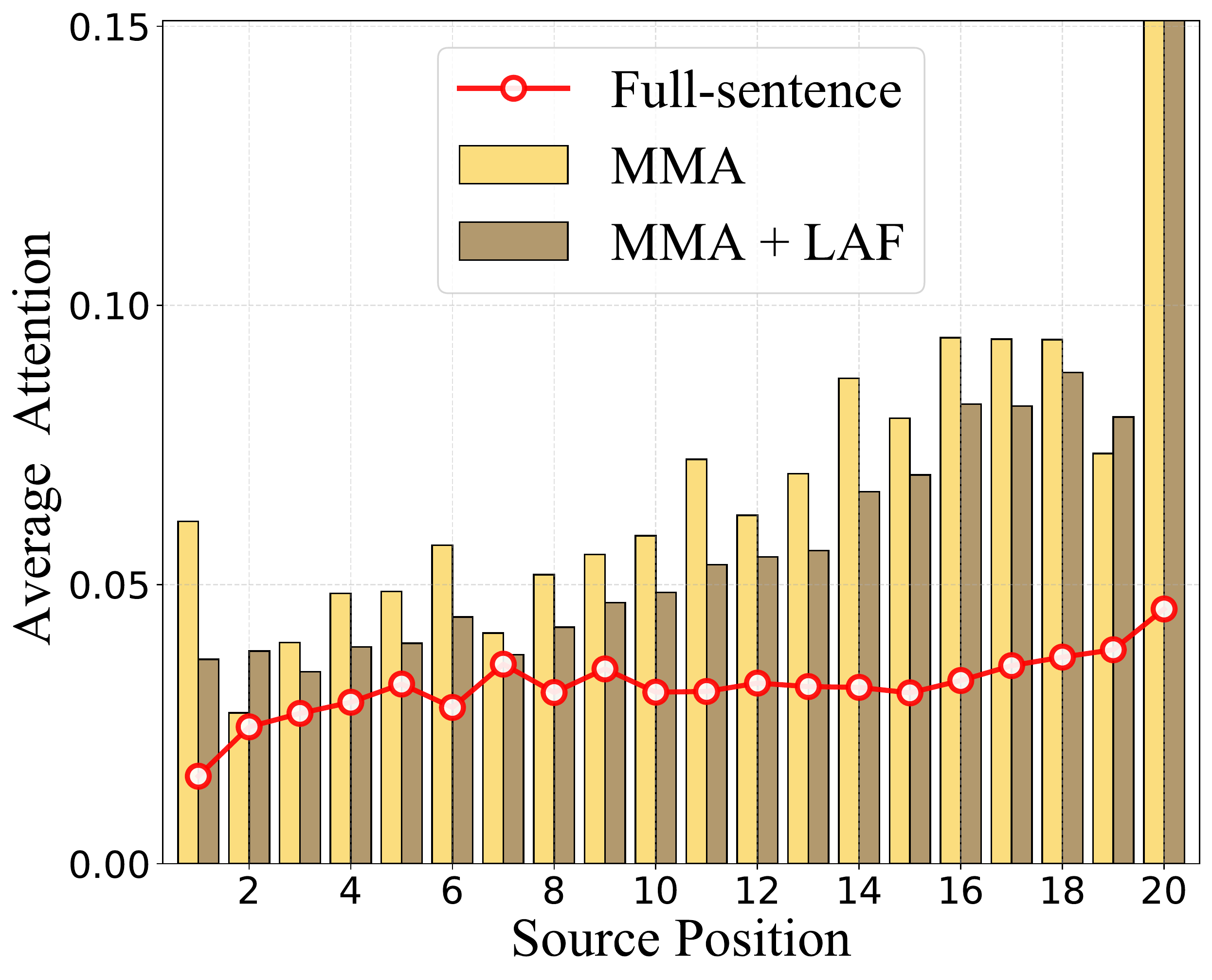}
\label{LAFpb2}
}

\caption{The improvements on average attention after applying LAF, where the position bias is reduced.}
\label{LAFpb}
\end{figure}

We show the change of average attention\footnote{Calculation is same with Eq.(\ref{eq5}) without calculating the future position predicted by LAF, so the comparison is fair.} after applying LAF in Figure \ref{LAFpb}. With LAF, the position bias in SiMT is significantly reduced, where the front positions are no longer illusoryly considered more important. By constructing the pseudo full-sentence, LAF bridges the structural gap between SiMT and full-sentence MT, so that the importance of source positions are more similar to that in full-sentence MT, thereby reducing the position bias. 

\subsection{Decreasing of Duplicate Translation}

\begin{figure*}[t]
\centering
\subfigure[Full-sentence MT]{
\includegraphics[width=1.23in]{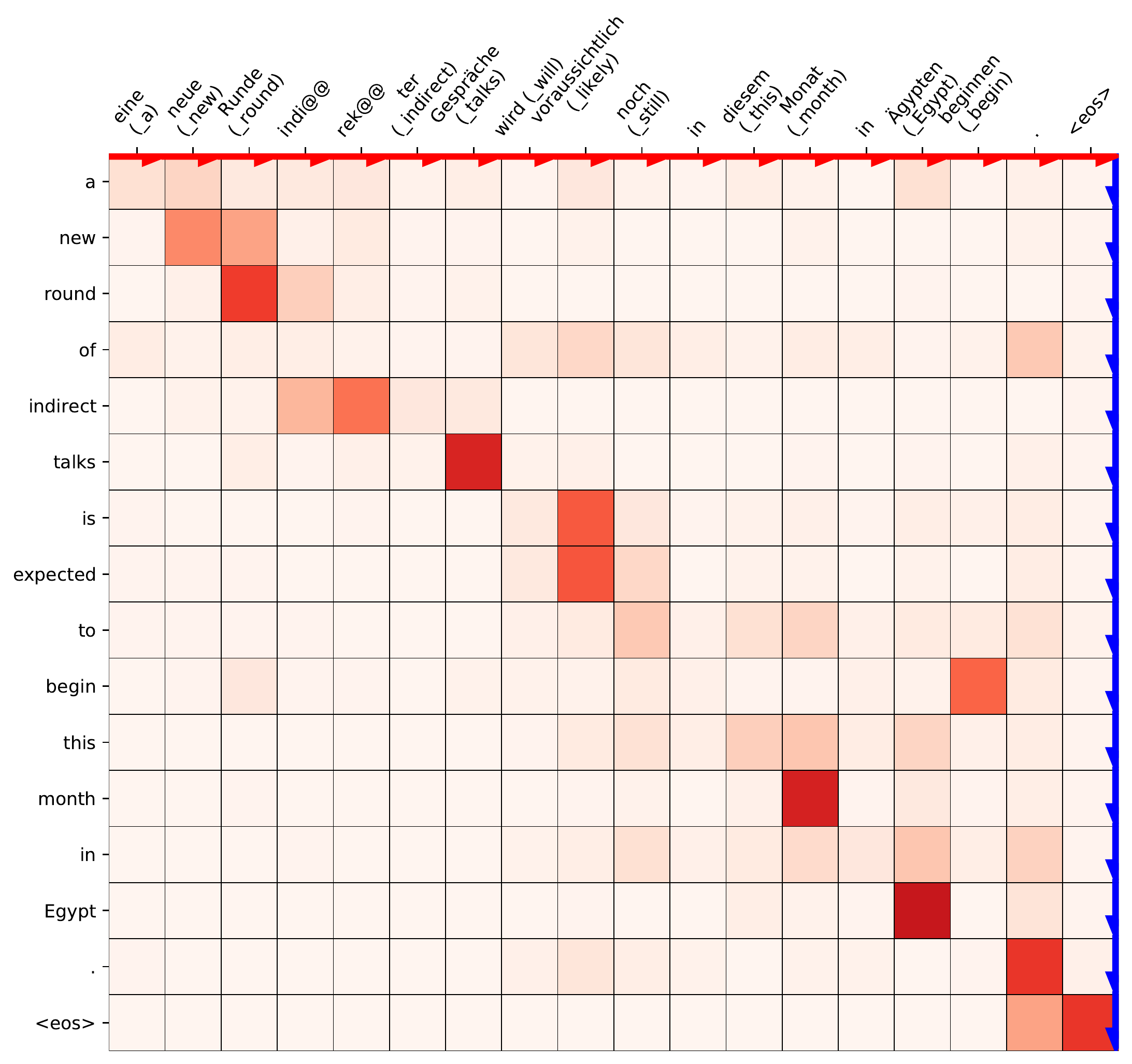}
\label{case1}
}\hspace{-3mm}
\subfigure[Wait-k]{
\includegraphics[width=1.23in]{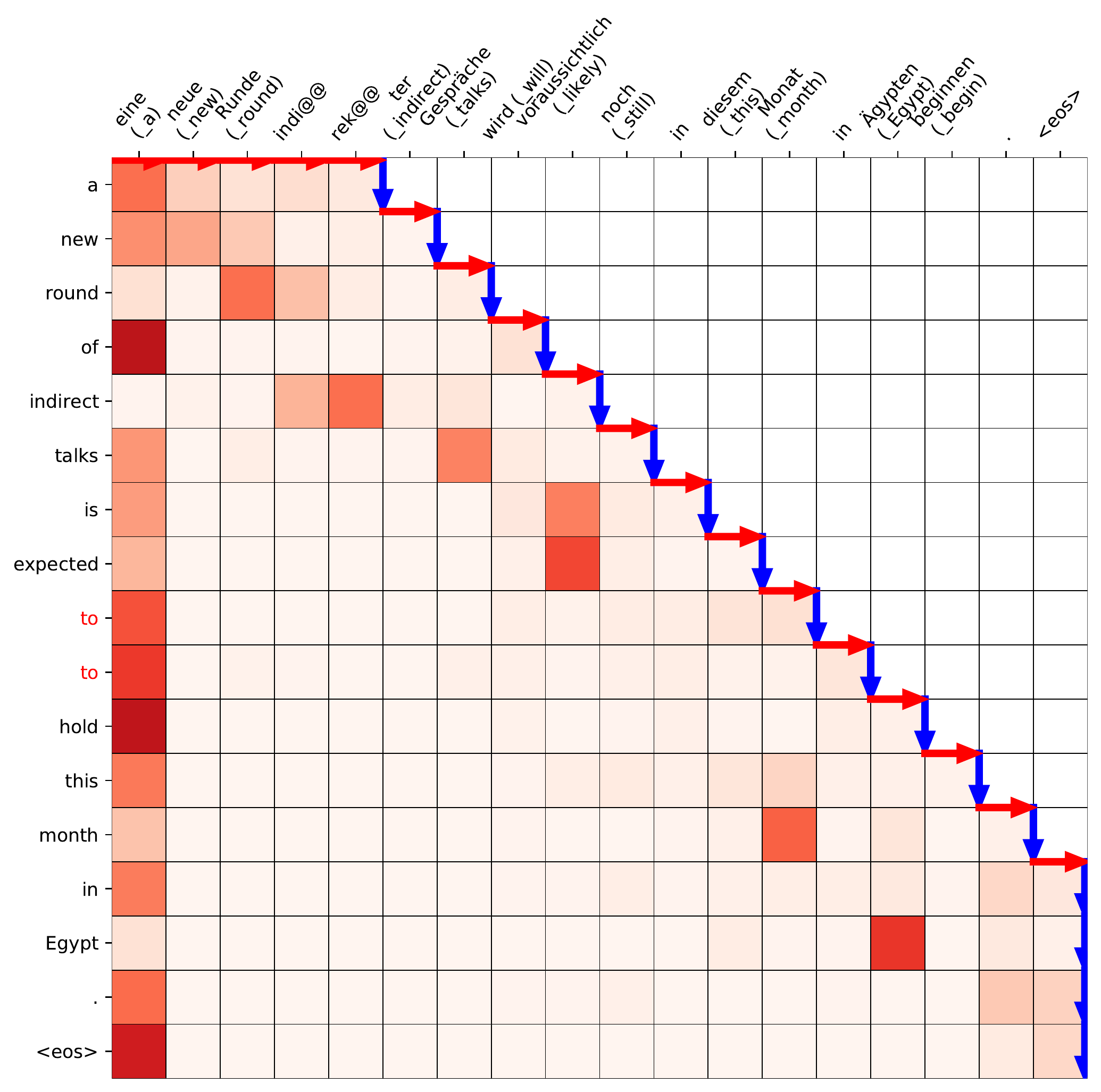}
\label{case2}
}\hspace{-3mm}
\subfigure[Wait-k + LAF]{
\includegraphics[width=1.23in]{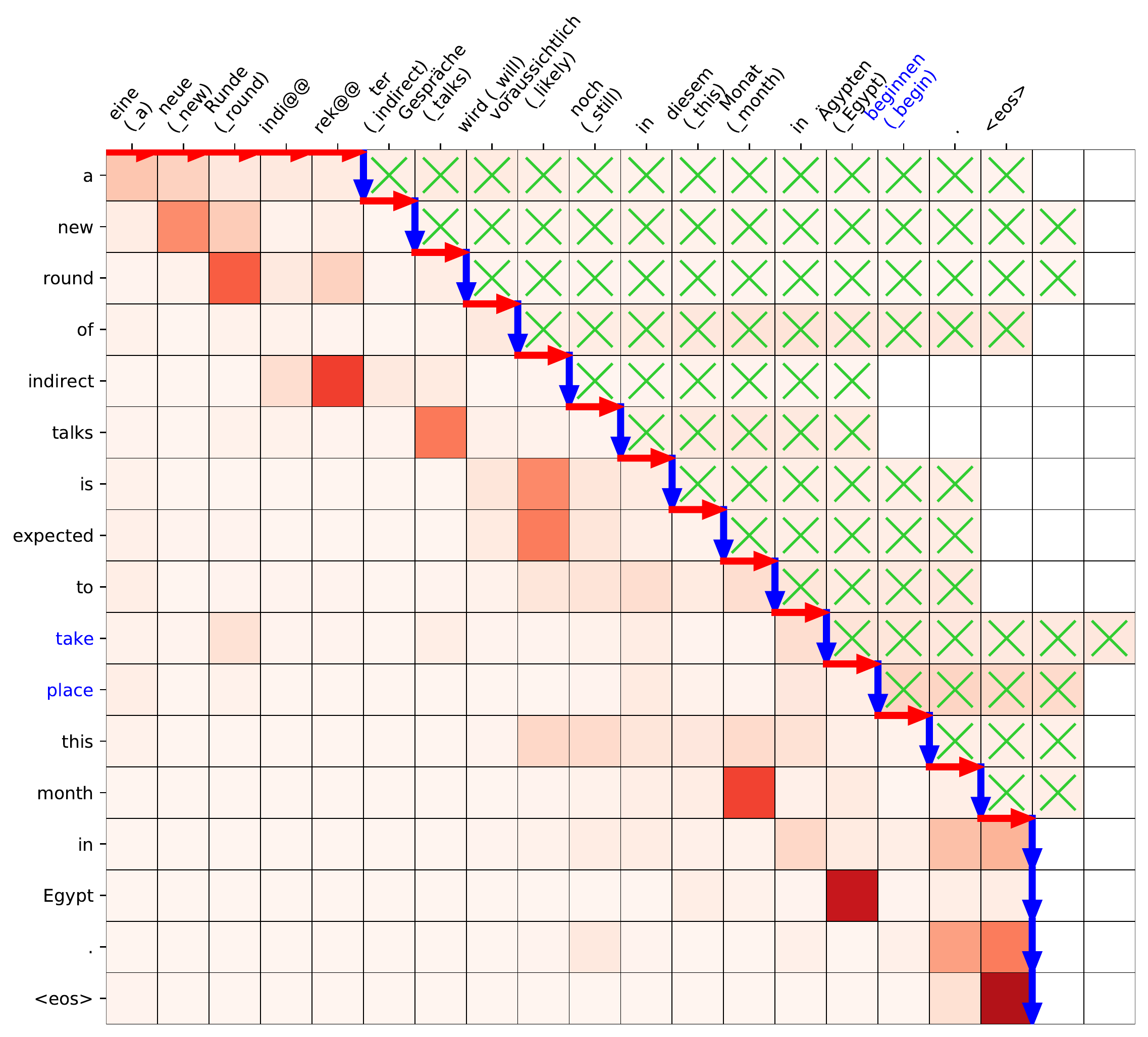}
\label{case3}
}\hspace{-3mm}
\subfigure[MMA]{
\includegraphics[width=1.23in]{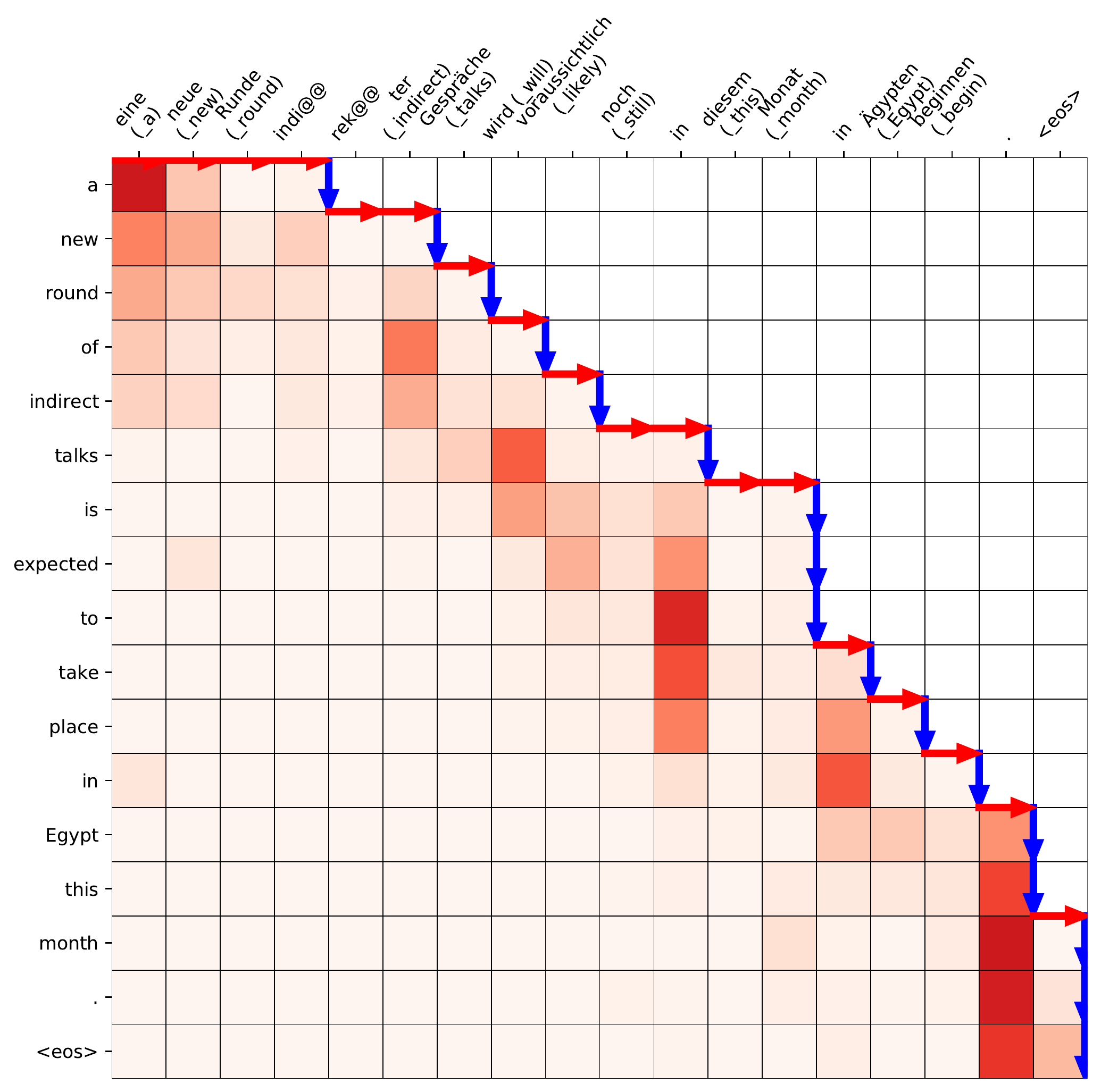}
\label{case4}
}\hspace{-3mm}
\subfigure[MMA + LAF]{
\includegraphics[width=1.23in]{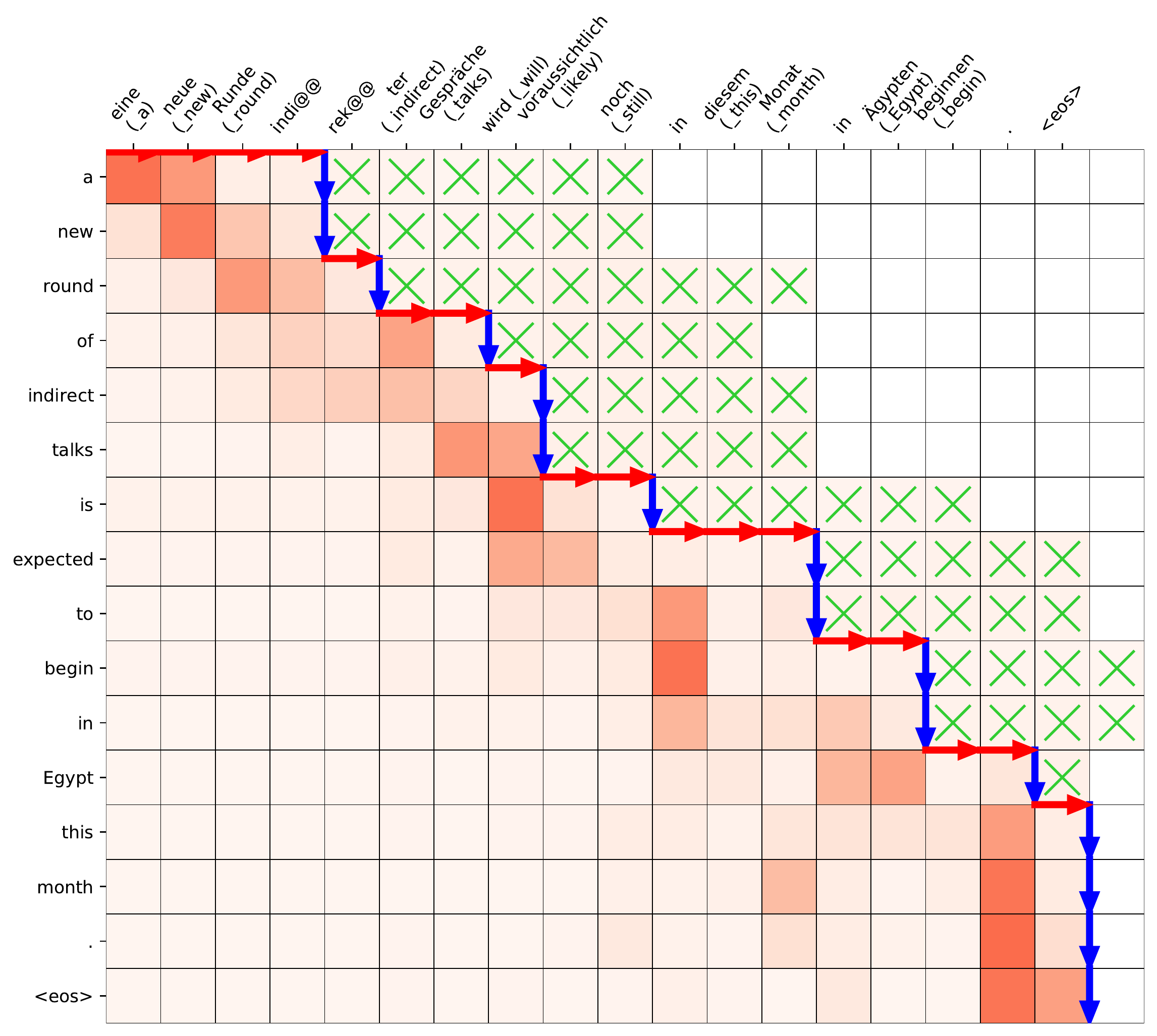}
\label{case5}
}
\caption{Attention visualization of a case on De$\rightarrow$En task. The horizontal axis is source input, and the vertical axis is target translation. The position with `\textcolor{green}{$\times $}' in LAF is the predicted future position filled with positional encoding. `\textcolor{red}{$\rightarrow$}': wait for a source word, `\textcolor{blue}{$\downarrow$}': translate a target word. The shade of the color indicates the attention weight.}
\label{case}
\end{figure*}

\begin{figure}[t]
\centering
\includegraphics[width=2.8in]{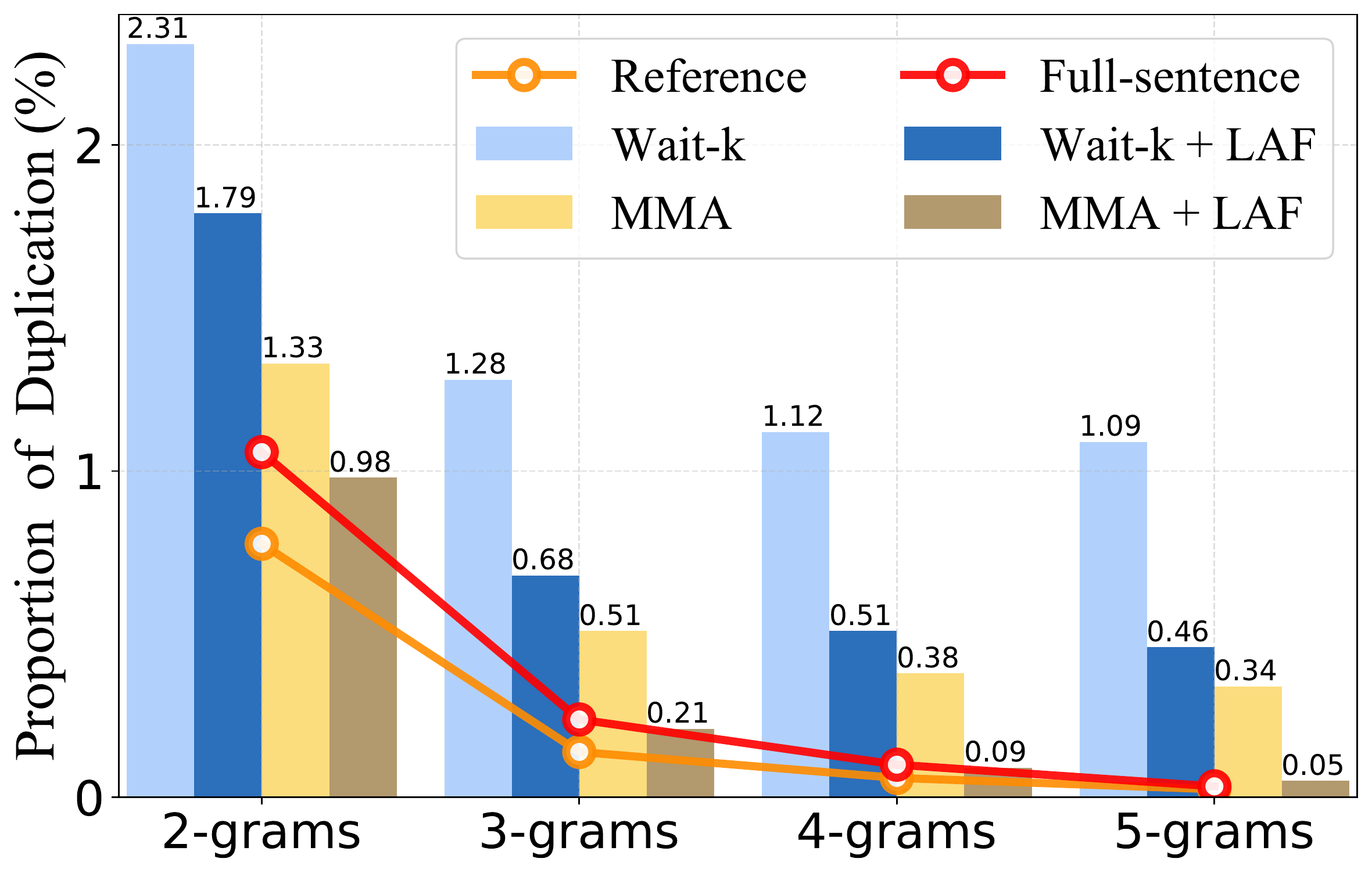}
\caption{Proportion of duplicate n-grams in translation, where LAF eliminates undesirable repetition.}
\label{du}
\end{figure}

Position bias makes the target word tend to focus on the front source word, which leads to much overlap in the attention distribution, resulting in duplicate translation errors \cite{elbayad-etal-2020-online}. Following \citet{see2017get}, we count the n-grams duplication proportion in translation in Figure \ref{du}.

There are few duplicate n-grams in reference and full-sentence MT, especially when $n\!>\!2$. However, position bias in SiMT makes the model always focus on some particular source words in the front position, thereby exacerbating duplicate translation errors, especially in the fixed policy. In 3-grams, the duplicate translation of Wait-k is about 6 times that of full-sentence MT, which is in line with the previous conclusion \cite{elbayad-etal-2020-online}. After applying LAF, the duplicate translation in SiMT is significantly reduced, similar to full-sentence MT.

\subsection{Improvement on Various Difficulty Levels}

\begin{table}[t]
\centering
\begin{tabular}{l|L{1.25cm}L{1.25cm}L{1.25cm}} \hline
              & $\,$\textbf{Easy}          & $\;$\textbf{Mid}           & \textbf{Hard}          \\ \hline
Full-sentence$\!\!\!$ & 34.32         & 31.93         & 30.91         \\ \hdashline
Wait-k         & 31.15         & 26.56         & 24.02         \\
Wait-k+LAF$\!\!\!$     & 32.93$^{+\!1.78}$         & 28.32$^{+\!1.76}$         &  \textbf{26.50}$^{\textbf{+\!2.48}}$         \\
MMA           & 29.17         & 26.94         & 25.09         \\
MMA+LAF$\!\!\!$       & 30.23$^{+\!1.06}$ & 27.99$^{+\!1.05}$ & \textbf{27.51}$^{\textbf{+\!2.42}}$ \\\hline
\end{tabular}
\caption{Improvement of our method on SiMT with various difficulty levels, which are divided according to the word order difference between the target and source.}
\label{diff}
\end{table}

The word order difference is a major challenge of SiMT, where many word order inversions may force the model to start translating before reading the aligned source words \cite{chen-etal-2021-improving-simultaneous}. Following \citet{zhang-feng-2021-universal}, We evenly divide the test set into three sets: Easy, Mid and Hard based on the number of reversed word orders in alignments using \texttt{fast-align}\footnote{\url{https://github.com/clab/fast_align}} \cite{dyer-etal-2013-simple}, and report the results on each set in Table \ref{diff}.

For full-sentence MT, word order reversal will not cause too much challenge, so that the performance gap between different sets is small. In SiMT, word order reversal often causes the model to translate before reading the aligned source words, which forces the target word to focus on some unrelated source words, resulting in poor performance in Hard set. LAF complements the incomplete source to the full-sentence length, which allows the target word to focus on the subsequent position instead of must focusing on the current irrelevant source word when the aligned word is not received, thereby obviously improving the performance on Hard set.

\subsection{Attention Characteristics}

\begin{figure}[t]
\centering
\includegraphics[width=2.9in]{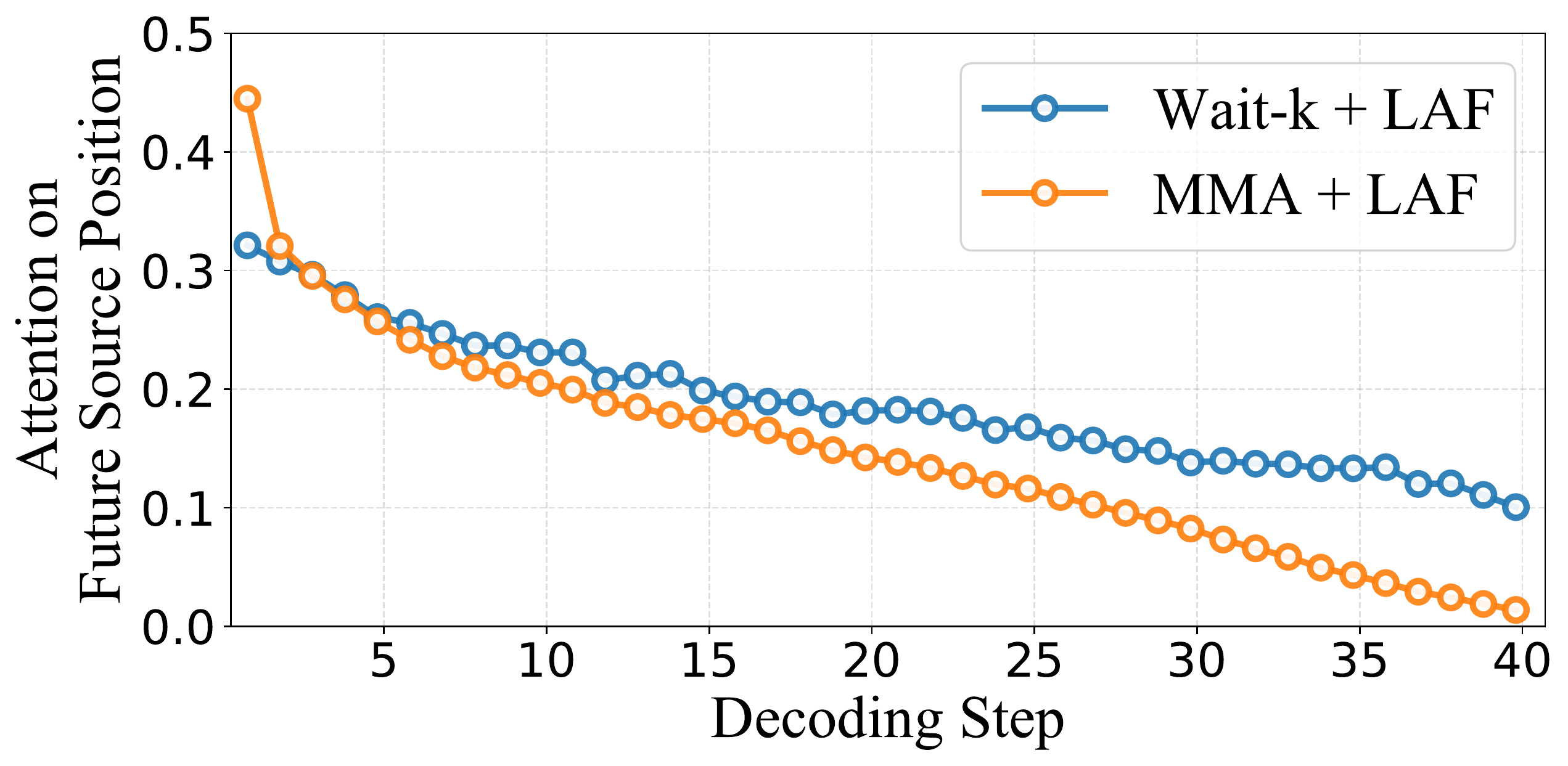}
\caption{The attention on future source position (filled with positional encoding) in different decoding steps.}
\label{pos_attn}
\end{figure}

LAF constructs the pseudo full-sentence by predicting the full-sentence length and filling the future position with positional encoding. To verify the importance of the future position, we count the attention weights on the future position (i.e., filled with positional encoding) at each decoding step in Figure \ref{pos_attn}. In the beginning, the future position gets much attention weight, especially getting about 30\% attention in the first decoding step. As the received source words increase, the attention received by future positions gradually decreases.


Furthermore, we visualize the attention distribution of an example in Figure \ref{case}. In Wait-k and MMA, attention is more concentrated on the front position, especially Wait-k extremely focuses on the first source word, which leads to duplicate translation ``\textit{expected to to hold}''. With LAF, when the aligned source word has not been received, the future positions tend to get more attention, e.g. when `Wait-k+LAF' translating ``\textit{take place}'' before receiving ``\textit{beginnen}''. Besides, the predicted length in LAF changes dynamically and gradually approaches the full-sentence length. Overall, LAF reduces the position bias and thus the attention in SiMT is more similar to the attention in full-sentence MT, resulting in better translation quality.

\section{Conclusion}
In this paper, we develop a length-aware framework for SiMT to reduce the position bias brought by incomplete source. Experiments show that our method achieves promising results by bridging the structural gap between SiMT and full-sentence MT.

\section*{Acknowledgements}
We thank all the anonymous reviewers for their insightful and valuable comments. This work was supported by National Key R\&D Program of China (NO. 2017YFE0192900).

\bibliography{anthology,custom}
\bibliographystyle{acl_natbib}

\newpage

\appendix
\onecolumn

\section{Theoretical Analysis of Position Bias in SiMT}

\label{sec:appendix}

In SiMT, each source position becomes unfair due to the streaming inputs, which leads to position bias. In this section, we conduct a theoretical analysis of position bias from the perspective of the difference in decoding probability.

\textbf{Full-sentence MT} We denote the source sentence as $\mathbf{x}\!=\!\left \{ x_{1},\cdots ,x_{J} \right \}$ with source length $J$, and target sentence as $\mathbf{y}\!=\!\left \{ y_{1},\cdots ,y_{I} \right \}$ with target length $I$. Given the source sentence $\mathbf{x}$, the decoding probability of full-sentence MT is calculated as:
\begin{align}
    p_{\!full}(\mathbf{y}\mid \mathbf{x})&=\prod_{i=1}^{I}p\left ( y_{i}\mid \mathbf{y}_{< i} , \mathbf{x} \right ) \\
    &=p\left ( y_{1}\mid \mathbf{x}\right )\times p\left ( y_{2}\mid y_{1}, \mathbf{x}\right )\cdots \times p\left ( y_{I}\mid y_{I-1}\cdots y_{1}, \mathbf{x} \right )\\
    &=\frac{p\left ( y_{1}, \mathbf{x} \right )}{p\left ( \mathbf{x} \right )}\times \frac{p\left ( y_{2},y_{1}, \mathbf{x} \right )}{p\left (  y_{1}, \mathbf{x} \right )}\cdots \times \frac{p\left ( y_{I}\cdots y_{1}, \mathbf{x} \right )}{p\left ( y_{I-1}\cdots y_{1}, \mathbf{x} \right )} \label{eq16}\\ 
    &= \frac{p(\mathbf{x},\mathbf{y})}{p(\mathbf{x})}\label{eq17}
\end{align}
where each target word $y_{i}$ is generated with complete $\mathbf{x}$, so that each source position is fair.

\textbf{Simultaneous machine translation} 
SiMT starts translating while receiving the streaming inputs and hence each target word is generated with a partial source prefix $\mathbf{x}_{\leq g(i)}$, where $g(i)$ is determined by a specific SiMT policy. Given the source sentence $\mathbf{x}$ and $g(i)$ (the number of received source words when generating $y_{i}$), the decoding probability of SiMT is calculated as:
\begin{align}
    p_{\!sim}(\mathbf{y}\mid \mathbf{x})&=\prod_{i=1}^{I}p\left ( y_{i}\mid \mathbf{y}_{< i} , \mathbf{x}_{\leq g(i)} \right ) \\
    &=p\left ( y_{1}\mid \mathbf{x}_{\leq g(1)} \right )\times p\left ( y_{2}\mid y_{1}, \mathbf{x}_{\leq g(2)} \right )\cdots \times p\left ( y_{I}\mid y_{I-1}\cdots y_{1}, \mathbf{x}_{\leq g(I)} \right ) \;\;\;  \\
    &=\frac{p\left ( y_{1}, \mathbf{x}_{\leq g(1)} \right )}{p\left ( \mathbf{x}_{\leq g(1)} \right )}\times \frac{p\left ( y_{2},y_{1}, \mathbf{x}_{\leq g(2)} \right )}{p\left (  y_{1}, \mathbf{x}_{\leq g(2)} \right )}\cdots \times \frac{p\left ( y_{I}\cdots y_{1}, \mathbf{x}_{\leq g(I)} \right )}{p\left ( y_{I-1}\cdots y_{1}, \mathbf{x}_{\leq g(I)} \right )} \label{eq20}
\end{align}
However, different from Eq.(\ref{eq16}) of full-sentence MT, the numerator and denominator of two adjacent items Eq.(\ref{eq20}) cannot be fully counteracted. Then, we decompose the denominator to counteract the numerator, and Eq.(\ref{eq20}) can be simplified as:
\begin{align}
    p_{\!sim}(\mathbf{y}\mid \mathbf{x})&=\frac{p\left ( y_{1}, \mathbf{x}_{\leq g(1)} \right )}{p\left ( \mathbf{x}_{\leq g(1)} \right )}\times \frac{p\left ( y_{2},y_{1}, \mathbf{x}_{\leq g(2)} \right )}{p\left (  y_{1}, \mathbf{x}_{\leq g(1)} \right )\times p\left ( _{g\left ( 1 \right )<  }\mathbf{x}_{\leq g\left ( 2 \right )} \mid y_{1}, \mathbf{x}_{\leq g\left ( 1 \right )} \right )} \times \cdots \nonumber\\  
    &\;\;\;\; \times \frac{p\left ( y_{I}\cdots y_{1}, \mathbf{x}_{\leq g(I)} \right )}{p\left ( y_{I-1}\cdots y_{1}, \mathbf{x}_{\leq g(I-1)} \right )\times p\left ( _{g\left ( I-1 \right )<  }\mathbf{x}_{\leq g\left ( I \right )} \mid y_{I-1}\cdots y_{1}, \mathbf{x}_{\leq g\left ( I-1 \right )} \right ) } \\
    &=\frac{p\left ( \mathbf{y},\mathbf{x}_{\leq g\left ( I \right )} \right )}{p\left ( \mathbf{x}_{\leq g(1)} \right )\times \prod_{i=2}^{I} p\left ( _{g\left ( i-1 \right )<  }\mathbf{x}_{\leq g\left ( i \right )} \mid \mathbf{y}_{< i}, \mathbf{x}_{\leq g\left ( i-1 \right )} \right )} \label{eq23}
\end{align}
where $_{g\left ( i-1 \right )<  }\mathbf{x}_{\leq g\left ( i \right )}$ represents the source words between $\left ( g\left ( i-1 \right ),g\left ( i \right ) \right ]$. Generally, the SiMT methods often ensure that in most cases the model has already received the complete source sentence before translating the last target word \cite{Arivazhagan2019,arthur-etal-2021-learning}, i.e. $\mathbf{x}_{\leq g\left ( I \right )}  \approx \mathbf{x}$. Therefore, Eq.(\ref{eq23}) can be written as:
\begin{align}
    p_{\!sim}(\mathbf{y}\mid \mathbf{x})=\frac{p\left ( \mathbf{x}, \mathbf{y}\right )}{p\left ( \mathbf{x}_{\leq g(1)} \right )\times \prod_{i=2}^{I} p\left ( _{g\left ( i-1 \right )<  }\mathbf{x}_{\leq g\left ( i \right )} \mid \mathbf{y}_{< i}, \mathbf{x}_{\leq g\left ( i-1 \right )} \right )}  \label{eq24}
\end{align}

\textbf{Comparison between SiMT and full-sentence MT} The decoding probability of full-sentence MT and SiMT are calculated as Eq.({\ref{eq17}}) and Eq.(\ref{eq24}), respectively. Compared with full-sentence MT, the streaming characteristics of SiMT reflects in the denominator of the decoding probability, which is no longer complete $\mathbf{x}$, but an autoregressive language model of $\mathbf{x}$. Therefore, SiMT needs to additionally model the sequential dependency of source sentence to predict next source segment $ _{g\left ( i-1 \right )<  }\mathbf{x}_{\leq g\left ( i \right )}$ based on previous source words $\mathbf{x}_{\leq g\left ( i-1 \right )}$ and target words $ \mathbf{y}_{< i}$. 

Due to the complexity and uncertainty of the sequential dependency between incomplete source words, it is difficult for SiMT to directly model the sequential dependency very well. Therefore, SiMT model always suffers from the issue of unfair source position caused by the sequential dependency, where the source words in the front position are illusoryly considered more important since the sequential dependency is left-to-right \cite{10.1162/tacl_a_00256,ZHANG2020103234,liu-etal-2021-scheduled-sampling}, resulting in the \emph{position bias}.

\textbf{Why length-aware framework work?} At each step $i$, given $\mathbf{x}_{\leq g\left ( i \right )}$, length-aware framework first predicts the full-sentence length and then fills the future source position with positional encoding, thereby turning the incomplete source words into pseudo full-sentence. 

Here, the predicted full-sentence length can be considered as a \emph{latent variable} during translating, aiming to help model the complex sequential dependency between incomplete source words, where introducing latent variable has been proven to provide effective help for modeling sequential dependency \cite{lee-etal-2018-deterministic,Su_Wu_Xiong_Lu_Han_Zhang_2018,Shu_Lee_Nakayama_Cho_2020,song-etal-2021-alignart}. 
Owing to the full-sentence length as the latent variable, the model has a stronger ability to model the sequential dependency, thereby reducing position bias.

\end{document}